\pdfoutput=1

\documentclass[11pt]{article}

\usepackage[preprint]{acl}

\usepackage{times}
\usepackage{latexsym}

\usepackage[T1]{fontenc}

\usepackage[utf8]{inputenc}

\usepackage{microtype}

\usepackage{inconsolata}

\usepackage{graphicx}
\usepackage{tcolorbox}

\tcbuselibrary{theorems}
\newtcbtheorem
  []
  {lesson}
  {Key Takeaway}
  {%
    theorem name,%
    colback=white,
    colframe=black!90,
    fonttitle=\bfseries,
  }
  {tay}

\usepackage{subfigure}

\newcommand{\noteng}[1]{\textcolor{red}{NG: #1}}
\newcommand{\notegb}[1]{\textcolor{blue}{GB: #1}}

\title{Evaluation of LLMs in Medical Text Summarization: The Role of Vocabulary Adaptation in High OOV Settings}

\author{
  \textbf{Gunjan Balde}\thanks{Corresponding Author.\\ Accepted for publication in the Findings of the 63rd Annual Meeting of the Association for Computational Linguistics (ACL 2025). This is the author’s version of the work. It is posted here for your personal use. Not for redistribution.},
  \textbf{Soumyadeep Roy},
  \textbf{Mainack Mondal}
  and \textbf{Niloy Ganguly}
\\
Indian Institute of Technology Kharagpur 
\\
\texttt{balde.gunjan0812@kgpian.iitkgp.ac.in}\\
\texttt{soumyadeep.roy9@iitkgp.ac.in} \\
\texttt{\{mainack,niloy\}@cse.iitkgp.ac.in}\\
}

\begin{document}
\maketitle
\begin{abstract}
Large Language Models (LLMs) recently achieved great success in medical text summarization by simply using in-context learning. However, these recent efforts do not perform fine-grained evaluations under difficult settings where LLMs might fail. They typically report performance scores over the entire dataset. Through our benchmarking study, we show that LLMs show a significant performance drop for data points with high concentration of out-of-vocabulary (OOV) words or with high novelty. Vocabulary adaptation is an intuitive solution to this vocabulary mismatch issue where the LLM vocabulary gets updated with certain expert domain (here, medical) words or subwords. An interesting finding from our study is that Llama-3.1, even with a vocabulary size of around 128K tokens, still faces \textit{over-fragmentation} issue with medical words. To that end, we show vocabulary adaptation helps improve the LLM summarization performance even in difficult settings. Through extensive experimentation of multiple vocabulary adaptation strategies, two continual pretraining strategies, and three benchmark medical summarization datasets, we gain valuable insights into the role of vocabulary adaptation strategies for customizing LLMs to the medical domain. We also performed a human evaluation study with medical experts where they found that vocabulary adaptation results in more relevant and faithful summaries. Our codebase is made publicly available at \url{https://github.com/gb-kgp/LLM-MedicalSummarization-Benchmark}.

\end{abstract}

\section{Introduction}



\begin{figure}[t]
    \centering
    \includegraphics[width=0.95\columnwidth]{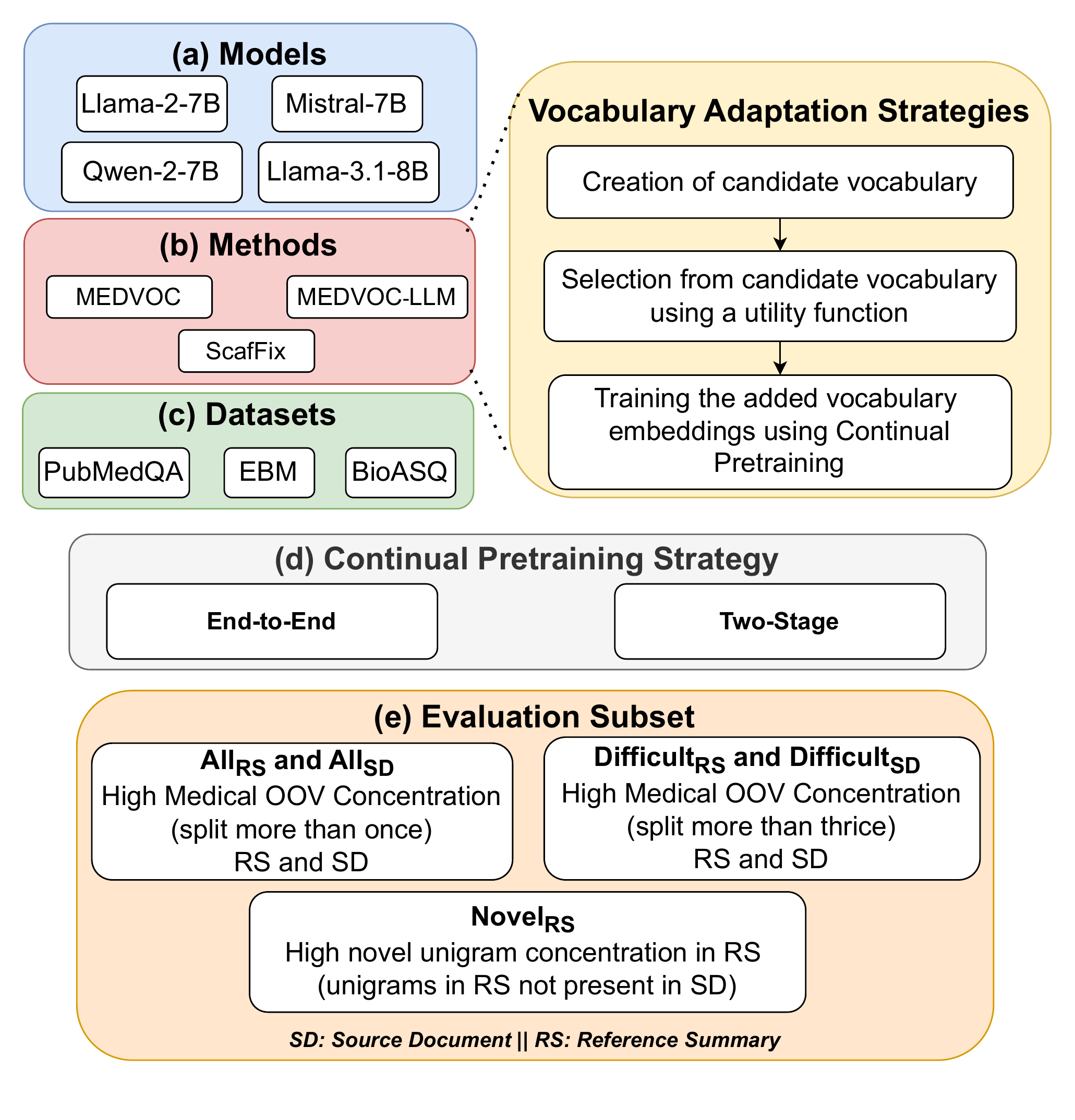}
    \caption{We present the benchmarking setup for our fine-grained evaluation of LLMs in high OOV and high novelty setting for medical text summarization. Our benchmarking setup: (a) tests four recent mainstream SoTA models; (b) on three vocabulary adaptation methods; (c) using three query-focused medical summarization dataset; (d) over two continual pretraining procedures; (e) evaluated in five fine-grained scenarios based on OOV (Out-Of-Vocabulary) and Novel unigram concentration in SD (Source Document) and RS (Reference Summary) respectively. So, in total we evaluated $4\times3\times3\times2\times5 = 360$ combinations. 
    }
    \label{fig:benchmark-setup}
\end{figure}

Recent works like Clinsumm~\cite{van2024adapted} explore various strategies to adapt LLMs in the task of medical text summarization. These strategies are (i) in-context learning~\cite{brown2020languagemodelsfewshotlearners,lampinen2022languagemodelslearnexplanations} where a fixed number of exemplars is added in the prompt at inference time only; and (ii) parameter-efficient fine-tuning using QLoRA~\cite{cite:QLoRA} because LLMs have billions of parameters which make complete finetuning computationally infeasible.  However, these studies do not perform fine-grained evaluations considering challenging generation scenarios and typically report performance scores over the entire dataset. In this paper, we consider two such challenging scenarios primarily related to the high vocabulary mismatch between LLM vocabulary and the medical domain. Figure~\ref{fig:benchmark-setup} provides an overview of our benchmarking setup. 

We note that LLMs during tokenization typically over-fragment (important) medical words, resulting in higher OOV concentration (proportion of words that are split into more than one token by an LLM tokenizer) compared to the generic open-domain text (e.g., news domain). In this work, we focus on the `BASE' model variants of four open-source, open-domain LLMs called Llama-2~\cite{touvron2023llama}, Mistal~\cite{jiang2023mistral}, Qwen2~\cite{yang2024qwen2technicalreport} and Llama-3.1~\cite{dubey2024llama}. Llama-3.1 and Qwen2 has a much larger vocabulary size of 128K and 151K tokens, which is significantly larger than Llama-2 and Mistral (almost four times), which have a vocabulary size of around 32K. Therefore, it would be interesting to investigate whether vocabulary adaptation still helps to improve Llama-3.1 and Qwen2. We observe in Table~\ref{tab:stats-frag-cnn-pac}, that there is an increase of $17.89\%$, $20.25\%$, $14.06\%$, and $13.08\%$ in fragment score~\cite{rust-etal-2021-good} when comparing tokenization of words from the open domain with the medical domain using Llama-2, Mistral, Qwen2, and Llama-3.1 tokenizers. This over-fragmentation affects the encoding stage as the semantic meaning is lost due to poor tokenization~\cite{hofmann-etal-2022-embarrassingly} and during generation, the model has to generate more subwords to generate a medical word. Examples of words that are over-fragmented is shown in Table~\ref{tab:example-medical-words}. 
Additionally, summarization datasets in the medical domain being from a specialized domain contain words in reference summaries (like disease names) which do not occur directly in the source document but require domain-knowledge to be inferred from the source document (novel words). 

\begin{table}[t]
\centering
\small
\begin{tabular}{lrr}
\hline
                       & \textbf{CNN-DM} & \textbf{PAC} \\ \hline
\multicolumn{3}{c}{\textbf{Llama-2}}                                  \\
Fragment Score         & 1.23             & \textbf{1.45}             \\
Split\textsubscript{$>3$} & 21\%             & \textbf{35\%}             \\ \hline
\multicolumn{3}{c}{\textbf{Mistral}}                                  \\
Fragment Score         & 1.17             & \textbf{1.41}             \\
Split\textsubscript{$>3$} & 21\%             & \textbf{34\%}             \\ \hline

\multicolumn{3}{c}{\textbf{Llama-3.1}}                       \\
Fragment Score         & 1.07             & \textbf{1.21}             \\ 
Split\textsubscript{$>3$} & 12\%             & \textbf{25\%}             \\ \hline
\multicolumn{3}{c}{\textbf{Qwen-2}}                       \\
Fragment Score         & 1.28             & \textbf{1.46}             \\ 
Split\textsubscript{$>3$} & 17\%             & \textbf{28\%}             \\
\hline     
\end{tabular}
\caption{OOV concentration and fragment score observed for general domain dataset (CNN-DailyMail) and medical domain dataset (PAC) obtained using tokenizers of models: Llama-2 and Mistral (Vocabulary size: 32K), and Llama-3.1 and Qwen-2 (Vocabulary size: 128K and 151K). Fragment score~\cite{rust-etal-2021-good} is the average number of subwords a word is tokenized into; and Split\textsubscript{$>3$} is the fraction of words split more than thrice. We note that for all the model tokenizers, irrespective of the vocabulary size, medical domain words are over-fragmented leading to higher fragment score (highlighted in bold), compared to general domain.}
\label{tab:stats-frag-cnn-pac}
\end{table}



Vocabulary expansion is a potential solution for such domain adaptation challenges that primarily arise due to vocabulary mismatch between the LLM vocabulary and the target expert-domain tasks. In this process, the model's vocabulary is expanded by incorporating tokens from the target domain that are not originally included in the main vocabulary. It has shown good performance for encoder-only models like BERT, RoBERTa on classification tasks~\cite{hofmann-etal-2022-embarrassingly,healthinf22,xu-etal-2023-retrieval} 
and encoder-decoder models like BART, PEGASUS, and Transformers-Large on summarization and machine translation tasks~\cite{xu-etal-2021-vocabulary,nag-etal-2023-entropy,balde2024ijcai,nag-etal-2024-cost}. Recent works are exploring vocabulary adaptation strategies for LLMs and proves useful for multilingual use-cases on models like Llama such as~\cite{liu-etal-2023-task,chinesellama,cui2024chatlawmultiagentcollaborativelegal,liu-etal-2024-gold,gao2024ve,nag-etal-2024-cost,tejaswi-etal-2024-exploring,yamaguchi2024can}. However, these studies primarily focus on languages other than English, which are relatively underrepresented in the pretraining corpus of LLMs.

We make the following contributions:
\begin{itemize}
    \item We investigate vocabulary adaptation strategies in using LLMs on tasks from the expert (medical) domain. Instead of evaluating a multilingual setting, we address a vocabulary mismatch within the same language (English). 
    \item We perform a benchmarking study on medical text summarization with various vocabulary adaptation strategies tailored for LLMs. We perform fine-grained evaluation of LLMs in high OOV and high novelty settings (Table~\ref{tab:results-settings}). 
    \item We conduct a human evaluation study with experts who also found that vocabulary adaptation produces more relevant, coherent and faithful summaries (Figure~\ref{fig:human-eval}).
\end{itemize}

\if{0}
\noteng{Have to discuss this and rewrite}
Our main findings are as follows --- 
\begin{itemize}
    \item We find that general-purpose LLMs over-fragment medical words more than the general domain words (Table~\ref{tab:stats-frag-cnn-pac}).
    \item Considering overall dataset, at least one vocabulary adaptation technique improves over or performs at par with the BASE model in 4/6 setting.
    \item Continual pretraining without vocabulary adaptation improves over BASE only in 4/12 high novelty settings and in 12/24 in high OOV settings.
    \item Vocabulary adaptation improves over BASE in high OOV concentrations in source documents and reference summaries in a total of 14/24 settings and in high novelty in a total of 7/12 settings. 

    \item Vocabulary adaptation techniques results in a drop in a fragment score (Table~\ref{tab:vocabulary-stats}) and improved word tokenization boundaries (Table.$X$) thus resulting in better performance. \notegb{Add reference to the table.}
\end{itemize}
\fi





\if{0}
\begin{itemize}
    \item Vocabulary adaptation improves over BASE LLMs overall only for the dataset with the least OOV concentration (Figure~\ref{fig:results-overall}). 
    \item Vocabulary adaptation helps in higher expert OOV concentrations in source documents and reference summaries in a total of 21 out of 24 settings.
    \item Vocabulary adaptation helps in cases when data points have higher novelty in a total of 4 out of 6 settings.
    \item Continual pretraining without vocabulary adaptation (CPT) improves only in higher novelty settings but not in higher OOV settings. Thus, highlighting the importance of vocabulary adaptation.
\end{itemize}
\fi


\section{Benchmarking Vocabulary Adaptation Strategies for LLMs}\label{sec:benchmark-base}
A vocabulary adaptation technique consists of three steps: (i) generating candidate vocabulary tokens from target downstream datasets, (ii) selecting important vocabulary tokens from candidate set using some utility function (e.g., fragment score~\cite{hong-etal-2021-avocado}, corpus entropy~\cite{xu-etal-2021-vocabulary}) to form added vocabulary,  and (iii) learning the embeddings of the added vocabulary and integration into LLM. In this work, we benchmarking the effect of all the three steps,combing the effect of first two steps: constructing the vocabulary and adding candidate tokens from the target domain to finalize the vocabulary for integration into the LLM as one, and then checking effect of last step: on how to train the embeddings.


\if{0}
\begin{figure}[t]
    \centering
    \includegraphics[width=\columnwidth]{figures/ACL-ARR-Feb15Cycle.pdf}
    \caption{Benchmarking setup of our fine-grained evaluation of LLMs in high OOV and high novelty setting for medical text summarization.} 
    \label{fig:benchmark-setup}
\end{figure}
\fi



\begin{table}[t]
\centering
\scalebox{0.9}{
\begin{tabular}{p{2.2cm}p{2.6cm}|p{2.3cm}}
\hline
 \textbf{Medical Term} & \textbf{Llama-2}                & \textbf{Llama-3.1}         \\ \hline
 \multicolumn{3}{c}{\textbf{High OOV}}             \\
cardiomyopathy & `\_card', `iom', `y', `op', `ath', `y' & `card', `i', `omy', `opathy' \\
antipyretics  & `\_ant', `ip', `y', `ret', `ics' & `ant', `ipy', `ret', `ics' \\ \hline
\multicolumn{3}{c}{\textbf{High novelty}}         \\
corticosteroid & `\_cort', `ic', `ost', `ero', `id'     & `c', `ortic', `oster', `oid'                     \\
antidepressant & `\_ant', `ide', `press', `ants'  & `ant', `idepress', `ants'  \\ \hline

\end{tabular}}
\caption{Medical terms from reference summary of PubMedQA dataset with high OOV concentration (\textit{Difficult\textsubscript{RS}}) and high novelty (\textit{Novel\textsubscript{RS}}).} 
\label{tab:example-medical-words}
\end{table}

\subsection{Vocabulary Adaptation Methods}\label{sec:methods-proposed-ScafFix}
We consider two types of datasets while constructing the vocabularies to be added to the PLM vocabulary: (i) PubMed Abstract Collection (PAC), a collection of around 300K PubMed abstracts used for intermediate fine-tuning by MEDVOC~\cite{balde2024ijcai}, and (ii) Target Task (TGT Task), the target downstream task dataset for which the vocabulary is to be constructed. We now describe the different vocabulary adaptation methods used for our benchmarking study. 

\paragraph{MEDVOC.} MEDVOC~\cite{balde2024ijcai} is a SoTA vocabulary adaptation strategy for adapting PLMs like BART and PEGASUS, on medical summarization tasks. First, candidate vocabularies are constructed on the medical OOV words from a domain-specific corpus (PAC) -- V\textsubscript{PAC}, and a downstream task dataset --V\textsubscript{TGT}. Then, an optimal vocabulary (V\textsubscript{MEDVOC}), that lies at the intersection of V\textsubscript{PAC} and V\textsubscript{TGT} is chosen via a hyperparameter search. The utility function for the hyperparameter search is fragment score~\cite{rust-etal-2021-good}, defined as the average number of subwords a word from a corpus $\mathcal{C}$ is tokenized into by a tokenizer using vocabulary $\mathcal{V}$. The vocabulary configuration within the neighborhood of the optimal vocabulary is finally chosen to avoid overfitting on large vocabulary sizes. 

\paragraph{MEDVOC-LLM.} This is a variant of MEDVOC adapted for LLM vocabularies and tokenizers. We identify two key issues in MEDVOC: (i) many of the vocabulary terms added directly from V\textsubscript{PAC},  did not occur even once in the reference summaries of the train set of downstream target task -- their addition did not contribute during generation, and (ii) certain added vocabulary terms were a mixture of numerals and punctuations (e.g., \emph{-9,}) -- which is not consistent \footnote{Llama tokenizers explicitly set apart digits as individual tokens~\cite{touvron2023llama}} with the tokenization scheme for LLMs considered in this study (Llama-2 and Llama-3.1). Thus, in this approach, we clean the vocabularies generated by MEDVOC by removing the terms from both categories and considering only the clean set.

\paragraph{Overhead in Previous Vocabulary Adaptation Strategy.} Consider the word ‘cholesterol’, which is not present in the Llama-3.1 model’s vocabulary and is tokenized by the Llama-3.1 tokenizer as \textit{[cho, le, sterol]}. Since a merge rule operates on pairs of tokens, we need to iteratively add pairs from left to right, as shown in Table~\ref{tab:iter_merge_rule_add}.

\begin{table}[h]
    \centering
    \begin{tabular}{ll}
    \hline
     \textbf{Token}    & \textbf{Merge Rule} \\ \hline
        `chole' & [ch, ole] \\
        `cholesterol' & [chole, sterol] \\ \hline

    \end{tabular}
    \caption{Illustration of iterative addition of tokens to add a target token `\textit{cholesterol}' in the vocabulary.}
    \label{tab:iter_merge_rule_add}
\end{table}

Thus, in order to add ‘cholesterol’ to the vocabulary, we need to add one extra token [chole]. Specifically, for Llama-3.1, we observe an overall addition of 20\% of extra such tokens, which are seldom used once the complete word is added into the vocabulary. This can lead to reduced performance in downstream tasks. To that end, we present ScafFix to address this issue with existing vocabulary adaptation strategies. 

\paragraph{ScafFix.} Unlike the vocabulary adaptation strategies described till now, ScafFix constructs the candidate set for added vocabulary by directly considering the medical words and ignores the tokenization step for forming candidate subwords. We directly select $x$ tokens (where $x$ represents the quota, set to 500 in this case, in steps of 50) from the candidate vocabulary tokens, ranking them by their frequency in descending order. We then follow the MEDVOC-based hyperparameter search optimizing fragment score to obtain the optimal vocabulary to be added. To offset the absence of such derivative tokens, we use AdaptBPE tokenization scheme~\cite{balde2024EMNLP} instead of the standard Llama tokenizers. Instead of directly utilizing merge rules, AdaptBPE first checks whether a part of the input token (using longest-first match) directly exists in the added vocabulary, preserves it from splitting, and then runs the merge-based byte-pair encoding scheme iteratively on the rest of the input. Here, we avoid including the scaffolding tokens~\cite{cognetta2024analysisbpevocabularytrimming,bauwens-delobelle-2024-bpe,chizhov-etal-2024-bpe,lian2024scaffoldbpe} during the vocabulary addition phase; these are derivative tokens that remain under-trained once the whole word is added to the vocabulary.

\if{0}
\paragraph{Overhead in addition of new tokens in the vocabulary.} One key observation from MEDVOC and MEDVOC\textsubscript{LLM} is that, due to the underlying byte-pair encoding mechanism, adding a new word to the vocabulary requires ensuring that the constituent subwords—which enable its construction through merging—are also present. 

 Such words are formally called, scaffolding terms~\cite{cognetta2024analysisbpevocabularytrimming,bauwens-delobelle-2024-bpe,chizhov-etal-2024-bpe,lian2024scaffoldbpe}, and end up being under-trained as they are barely utilized independently during the tokenization phase. 
\fi

\subsection{Learning the Added Vocabulary Embeddings using Continued Pretraining}\label{sec:train-vocab-methods}
The embedding of the newly added vocabulary is initialized as the average of the embeddings of the existing subwords in the vocabulary~\cite{yamaguchi2024can}. We explore two continual pretraining~\cite{gururangan-etal-2020-dont,tejaswi-etal-2024-exploring,yamaguchi2024can} strategies to train these embeddings on target domain text (20K random documents from the PubMed Abstract Collection, in our case). Continual pretraining uses the same training objective of `Next Token Prediction' as the autoregressive language modeling objective, and optimizes the standard negative log-likelihood loss. We use the popular parameter-efficient fine-tuning technique known as `Low-Rank Adaptors' (LoRA)~\cite{peft} because end-to-end training of LLMs is computationally infeasible.
We explore two continual pretraining strategies:  
\begin{itemize}
    \item \textbf{End-To-End}: The model is trained in an end-to-end manner by freezing all the base model layers except the input and output embedding layers and training LoRA adapters.
    \item \textbf{Two-Stage}: First, the entire model is frozen along with the LoRA layers except for the input and output embedding layers for a short duration, then unfreeze the LoRA adapters and train the LoRA adapters along with the embedding layers~\cite{chinesellama,yamaguchi2024can}. The second approach leads to more stable training and avoids overfitting to the initial LLM embedding space.
\end{itemize}

\if{0}
\begin{equation} \label{eq-NLL}
\mathcal{L}(\theta) = - \sum_{t=1}^{T} \log P_{\theta}(x_t \mid x_{<t}),
\end{equation}
\fi

\section{Experimental Setup}
We use in-context learning (ICL)~\cite{brown2020languagemodelsfewshotlearners,van2024adapted} along with greedy decoding for generating summaries from LLMs. We follow the approach similar to ClinSumm~\cite{van2024adapted}, where examples are sampled from the train set using the similarity computed using PubMedBERT\footnote{\url{https://huggingface.co/pritamdeka/PubMedBERT-mnli-snli-scinli-scitail-mednli-stsb}} with the given test data point. The template for prompting is shown in Appendix~\ref{appendix:expt-setup} (Table~\ref{appendix:tab-prompt}). We also describe the fine-grained evaluation setup used for this benchmarking study in Table~\ref{tab:definition-subset}.

\begin{table}[t]
\centering
\scalebox{0.87}{
\begin{tabular}{lp{5.5cm}}
\hline
\multicolumn{1}{c}{\textbf{Category}} & \multicolumn{1}{c}{\textbf{Description}}                                           \\ \hline
\multicolumn{2}{c}{\textbf{Categories of Words}}                                                     \\                              
Difficult-OOV                        & Medical words that are split more than thrice by model tokenizers        \\
Novel                          & Words in the summary that are not present in the source document                   \\ 
All-OOV                         & Medical words that are split more than once by model tokenizers             \\ \hline
\multicolumn{2}{c}{\textbf{Evaluation Setting (Top ten percentile)}}                                                                  \\ 
Difficult\textsubscript{RS}                          & High Difficult-OOV concentration in the reference summaries \\
Difficult\textsubscript{SD}                          & High Difficult-OOV concentration in the source document     \\
Novel\textsubscript{RS}                              & High Novel concentration in the reference summaries  \\ 
All\textsubscript{SD}                                & High All-OOV concentration in the source document      \\
All\textsubscript{RS}                                & High All-OOV concentration in the reference summaries  \\ \hline

\end{tabular}}
\caption{Challenging fine-grained evaluation scenarios considered in this study. We focus on the subset that contains high OOV concentration and novelty}
\label{tab:definition-subset}
\end{table}

Next, we describe the benchmark medical text summarization datasets used, followed by details on evaluation metrics, baseline models, and training details. 

\subsection{Datasets} 
In this work, we focus on three query-focused medical summarization datasets such as BioASQ~\cite{tsatsaronis2015overview}, EBM~\cite{molla2011development}, and PubMedQA~\cite{jin2019pubmedqa}. A data point consists of a Query (Q) and a context (PubMed Abstract) as the input to the model. The gold-standard reference summary (RS) summarizes the abstract based on the query. We present the overall dataset characteristics in Table~\ref{tab:dataset-stats}.

\begin{table*}[t]
\centering
\scalebox{0.73}{
\begin{tabular}{lccrcrrrcrrrc}
\hline 
\multicolumn{1}{c}{\textbf{Dataset}} &
  \textbf{\begin{tabular}[c]{@{}c@{}}Test Set \\ Size\end{tabular}} &
  \multicolumn{2}{c}{\textbf{\begin{tabular}[c]{@{}c@{}}Token Count of\\ Reference Summaries\end{tabular}}} &
  \multicolumn{4}{c}{\textbf{\begin{tabular}[c]{@{}c@{}}OOV Concentration\\ Split more than once (in \%)\end{tabular}}} &
  \multicolumn{4}{c}{\textbf{\begin{tabular}[c]{@{}c@{}}OOV Concentration\\ Split more than thrice (in \%)\end{tabular}}} &
  \textbf{\begin{tabular}[c]{@{}c@{}}Unigram Novelty\\ (in \%)\end{tabular}} \\ 
\multicolumn{1}{c}{} & & \multicolumn{1}{r}{\textbf{Llama-2}} & \multicolumn{1}{r}{\textbf{Llama-3.1}} & \multicolumn{2}{c}{\textbf{Llama-2}} & \multicolumn{2}{c}{\textbf{Llama-3.1}} & \multicolumn{2}{c}{\textbf{Llama-2}} & \multicolumn{2}{c}{\textbf{Llama-3.1}} & \\
\multicolumn{1}{c}{} & & \textbf{} & \multicolumn{1}{c}{\textbf{}} &
\textbf{SD} & \multicolumn{1}{c}{\textbf{RS}} & \multicolumn{1}{c}{\textbf{SD}} & \multicolumn{1}{c}{\textbf{RS}} & \textbf{SD} & \multicolumn{1}{c}{\textbf{RS}} & \multicolumn{1}{c}{\textbf{SD}} & \multicolumn{1}{c}{\textbf{RS}} &  \\ \hline
PubMedQA & \multicolumn{1}{r}{500} & \multicolumn{1}{r}{63} & 25 & \multicolumn{1}{r}{36.67} & \multicolumn{1}{r}{38.00} & 43.68 & 45.65 & \multicolumn{1}{r}{4.91} & \multicolumn{1}{r}{4.65} & 2.61 & 2.42 & \multicolumn{1}{r}{41.32} \\ 
EBM & \multicolumn{1}{r}{424} & \multicolumn{1}{r}{112} & 91 & \multicolumn{1}{r}{38.97} & \multicolumn{1}{r}{40.90} & 45.60 & 46.23 & \multicolumn{1}{r}{6.65} & \multicolumn{1}{r}{7.92} & 3.90 & 5.17 & \multicolumn{1}{r}{47.15} \\
BioASQ-$\mathcal{M}$ & \multicolumn{1}{r}{963} & \multicolumn{1}{r}{85} & 69 & \multicolumn{1}{r}{46.20} & \multicolumn{1}{r}{50.64} & 52.03 & 56.61 & \multicolumn{1}{r}{9.12} & \multicolumn{1}{r}{11.04} & 5.55 & 7.09 & \multicolumn{1}{r}{42.58} \\
BioASQ-$\mathcal{S}$ & \multicolumn{1}{r}{496} & \multicolumn{1}{r}{73} & 58 & \multicolumn{1}{r}{47.12} & \multicolumn{1}{r}{50.00} & 52.00 & 57.15 & \multicolumn{1}{r}{8.70} & \multicolumn{1}{r}{9.10} & 4.76 & 4.55 & \multicolumn{1}{r}{4.11} \\ \hline
\end{tabular}}
\caption{Medical text summarization datasets used for evaluation. We have three key observations: (i) BioASQ has the highest OOV concentration; (ii) EBM has highest novelty concentration; and (iii) EBM has the longest length reference summaries.} 
\label{tab:dataset-stats}
\end{table*}

\paragraph{PubMedQA.} PubMedQA is a question-answering dataset with 1000 human-annotated data points. Here, input is a query appended to a PubMed abstract, which forms the source document. We consider the `long answer' as the reference summaries. 

\paragraph{EBM.} In this single-document summarization task, the input comprises a query paired with a PubMed abstract. The task is to generate a concise summary that addresses the query, using the provided abstract as its context.

\paragraph{BioASQ.} We use the Phase-B query-focused summarization task of BioASQ-9B. The input includes a question along with relevant PubMed abstracts. For the summarization task, we use the ideal answer as the reference summary. Here, we explore two variants of a source document --- (i) Snippets (BioASQ-$\mathcal{S}$): the question followed by the list of relevant snippets from a collection of PubMed Abstracts, in line with the MEDVOC paper~\cite{balde2024ijcai}, and (ii) BioASQ-Main Abstract (BioASQ-$\mathcal{M}$): the question followed by the complete Pubmed abstracts. 


\if{0}
\subsection{Fine-grained Evaluation Settings}
We evaluate the generalization capabilities of LLMs in fine grained settings of high OOV (Out-Of-Vocabulary) concentrations and novelty. We consider two categories of such OOVs: (i) \emph{All-OOV} -- medical terms that are split more than once by the LLM tokenizer (e.g. surgery: {\it [\textbf{\_}surg, ery]}), and (ii) \emph{Difficult-OOV} -- medical terms that are split more than thrice by the LLM tokenizer (e.g. antioxidant: {\it [\_ant, io, x, id, ant]}). The latter category terms are much more difficult to generate (and encode) as the semantic meaning of such terms is lost due to excessive tokenization~\cite{hofmann-etal-2022-embarrassingly}. Additionally, we also evaluate the cases when the unigrams present in the reference summaries are not directly present in the input source document --\emph{Novel-RS}. During the evaluation, we consider only the top ten percentile of data points which present high concentrations of OOVs and novelty considering the source document (Difficult\textsubscript{SD} and All\textsubscript{SD}) and reference summary (Difficult\textsubscript{RS} and All\textsubscript{RS}). The fine-grained evaluation settings are summarized in Table~\ref{tab:definition-subset}.
\fi

\subsection{Evaluation Metrics}
We evaluate the model-generated summaries using Rouge-L (R-L) to measure informativeness and coherence, and Concept-Score (CSr) to measure faithfulness~\cite{zhang2023famesumm}. Concept-Score measures the overlap of UMLS medical concepts (computed using QuickUMLS tool~\cite{soldaini2016quickumls}) between the generated and reference summaries. We use Rouge-L as the primary comparison metric, in line with prior studies~\cite{fabbri-etal-2021-summeval,yuan-etal-2022-biobart,balde2024EMNLP, balde2024ijcai}.


\subsection{Baseline Models}
The following baseline models do not update the LLM vocabulary:

\begin{itemize}
    \item \textbf{BASE.} `BASE' corresponds to the base variant of original LLM without any vocabulary adaptation and continual pretraining. We consider 7B variant of Llama-2 (Model id: \href{https://huggingface.co/meta-llama/Llama-2-7b-hf}{meta-llama/Llama-2-7b-hf}), Mistral (Model id: \href{https://huggingface.co/mistralai/Mistral-7B-v0.1}{mistralai/Mistral-7B-v0.1}), Qwen-2 (Model id: \href{https://huggingface.co/Qwen/Qwen2-7B}{Qwen/Qwen2-7B}), and 8B variant of Llama-3.1 (Model id: \href{https://huggingface.co/meta-llama/Llama-3.1-8B}{meta-llama/Llama-3.1-8B}) as our BASE models.

    \item \textbf{Continual Pretraining (CPT-Only).} CPT-Only corresponds to the original BASE LLM that has undergone the continual pretraining (CPT) but without any vocabulary adaptation. It serves as a strong baseline when comparing against vocabulary-adapted models.
\end{itemize}

\subsection{Implementation Details}
We provide basic details of implementation in terms of its LoRA implementation, pretraining corpus and training hyperparameters. 

\paragraph{Continual Pretraining using LoRA.} We use one A100 40 GB GPU to carry out the pretraining. We use LoRA~\cite{hu2022lora,peft} to carry out the pretraining in this resource-constrained setting. The LoRA adapters are applied to all the linear modules in the model. These include $\{k\_proj,\text{ } q\_proj,\text{ } v\_proj, \text{ and } o\_proj\}$ modules from self attention layers along with $\{gate\_proj,\text{ } up\_proj, \text{ and } down\_proj\}$ modules from MLP layers. We use a consistent LoRA configuration of a rank value of 32 and an alpha value of 64 for fair comparison. 

\paragraph{Size of Pretraining Corpus.} As continual pretraining with all the 312K documents from the PubMed Abstracts Collection (PAC) is computationally infeasible for LLMs, we perform a hyperparameter search to identify an optimal dataset size. We experiment with various dataset sizes of 10K, 20K, 50K, and 100K, and observe that the performance observed for 20K which only takes 6 hours of pretraining was comparable with the of 100K setting, which took approximately 40 hrs of pretraining. Therefore, we \textit{continually pretrain over the BASE model with randomly selected 20K documents from PAC}.


\paragraph{Training Hyperparameters.} We use a global batch size of 32 (on device: 8 with gradient accumulation: 4), and a learning rate of $1e-4$. In End-to-End pretraining procedure, LoRA layers of the model were trained end-to-end for 5 epochs. In case of Two-Stage pretraining procedure, the embedding layers training phase and keeping LoRA layers frozen, was carried out for 2 epochs on 10K PAC samples. Then both the LoRA and embedding layers were trained end-to-end on 20K PAC samples for 3 epochs. In both the pretraining procedures, the base layers of the models were kept frozen throughout the training process.

\begin{table*}[!ht]
\centering
\scalebox{0.68}{
\begin{tabular}{llllll|lllll}
\hline
 & \multicolumn{5}{c}{\textbf{Llama-2}} & \multicolumn{5}{c}{\textbf{Llama-3.1}} \\
\textbf{Model} & \textbf{Test\textsubscript{Full}} & \textbf{Difficult\textsubscript{RS}} & \textbf{Novel\textsubscript{RS}} & \textbf{Difficult\textsubscript{SD}} & \textbf{Overall} & \textbf{Test\textsubscript{Full}} & \textbf{Difficult\textsubscript{RS}} & \textbf{Novel\textsubscript{RS}} & \textbf{Difficult\textsubscript{SD}} & \textbf{Overall} \\ \hline
\multicolumn{11}{c}{\textbf{PubMedQA}} \\ 
BASE & 26.33 & 24.40 & 19.78 & 23.53 & 23.51 & \textbf{28.10} & 26.87 & 18.87 & 24.46 & 24.58 \\
CPT-only & 27.12 & 28.00 & 22.21 & 25.97 & 25.83 & 26.62 & \textbf{27.08} & 21.07 & 24.52 & 24.82 \\
MEDVOC & 26.50 & 22.88 & 20.14 & 24.35 & 23.47 & 27.86 & 26.23 & 21.87 & 24.68 & 25.16 \\
MEDVOC-LLM & 26.90 & 26.67 & 20.00 & 24.49 & 24.52 & 27.69 & 26.67 & 21.74 & \textbf{26.51} & \textbf{25.65} \\
ScafFix & \textbf{27.61} & \textbf{28.57} & \textbf{22.22} & \textbf{26.32} & \textbf{26.18} & 27.67 & 25.00 & \textbf{22.22} & 23.85 & 24.69 \\ \hline 

\multicolumn{11}{c}{\textbf{EBM}} \\ 
BASE & 18.56 & \textbf{16.00} & 11.20 & 17.33 & 15.77 & 20.04 & 14.54 & 11.37 & 15.79 & 15.44 \\
CPT-only & 19.13 & 15.87 & 11.77 & \textbf{19.05} & \textbf{16.46} & 20.13 & 17.15 & 13.33 & 16.98 & 16.90 \\
MEDVOC & 18.60 & 15.38 & 12.50 & 18.65 & 16.28 & 20.31 & 16.03 & 12.17 & 17.93 & 16.61 \\
MEDVOC-LLM & \textbf{19.27} & 14.71 & \textbf{12.90} & 17.54 & 16.11 & 20.75 & 16.40 & 13.04 & \textbf{18.39} & \textbf{17.15} \\
ScafFix & 18.65 & 14.71 & 11.77 & 17.40 & 15.63 & \textbf{20.79} & \textbf{17.17} & \textbf{13.80} & 16.18 & 16.99 \\ \hline 

\multicolumn{11}{c}{\textbf{BioASQ-$\mathcal{S}$}} \\
BASE & 32.12 & 26.12 & 20.53 & 35.72 & 28.62 & 35.25 & 29.60 & 18.18 & 27.03 & 27.52 \\
CPT-only & \textbf{33.30} & 29.22 & 20.00 & 30.33 & 28.21 & 36.01 & 24.70 & 17.14 & 29.41 & 26.82 \\
MEDVOC & 32.26 & 28.17 & 18.18 & 27.40 & 26.50 & 37.01 & 29.79 & 18.18 & 32.43 & 29.35 \\
MEDVOC-LLM & 32.40 & \textbf{29.41} & 20.29 & 30.33 & 28.11 & \textbf{37.15} & \textbf{32.89} & 20.29 & \textbf{37.84} & \textbf{32.04} \\
ScafFix & 32.88 & 29.29 & \textbf{22.22} & \textbf{36.44} & \textbf{30.21} & 36.70 & 32.26 & \textbf{20.90} & 34.15 & 31.00 \\ \hline

\multicolumn{11}{c}{\textbf{BioASQ-$\mathcal{M}$}} \\

BASE 		& \textbf{28.50} & 27.27 & \textbf{21.53} & 28.00 & 26.33 & \textbf{29.28} & 26.67 & \textbf{21.51} & 28.57 & 26.51  \\
CPT-only 	& 27.22 & 27.91 & 20.31 & 28.57 & 26.00 & 27.56 & 26.09 & 19.84 & 28.57 & 25.52  \\
MEDVOC 		& 24.19 & 23.26 & 16.53  & 26.32 & 22.73 & 27.71 & 27.27 & 20.46 & 28.57 & 26.01  \\
MEDVOC-LLM 	& 24.50 & 25.00 & 18.90 & 28.00  & 24.10 & 27.45 & 27.27 & 19.05 & 29.17 &  25.74 \\
ScafFix 	& 26.16 & \textbf{29.79} & 21.17 & \textbf{30.00} & \textbf{26.78} & 28.91 & \textbf{27.59} & 21.23  & \textbf{30.00} & \textbf{26.94} \\ \hline
\end{tabular}}
\caption{Fine-grained performance evaluation of vocabulary adaptation strategies and baseline modes in terms of Rouge-L (\textbf{R-L}). The values represent the best among the two continual pretraining strategies of `End-to-End' and `Two-Stage'. The `Overall' column represents the average value of Test\textsubscript{Full}, Difficult\textsubscript{RS}, Novel\textsubscript{RS} and Difficult\textsubscript{SD}. Vocabulary adaptation (best of MEDVOC-LLM and ScafFix) improves over BASE and CPT-only in $7$ out of $8$ overall settings.} 
\label{tab:results-settings}
\end{table*}

\section{Experimental Results}
\begin{figure*}[t]
    \centering
    \subfigure[]{\includegraphics[width=0.32\textwidth]{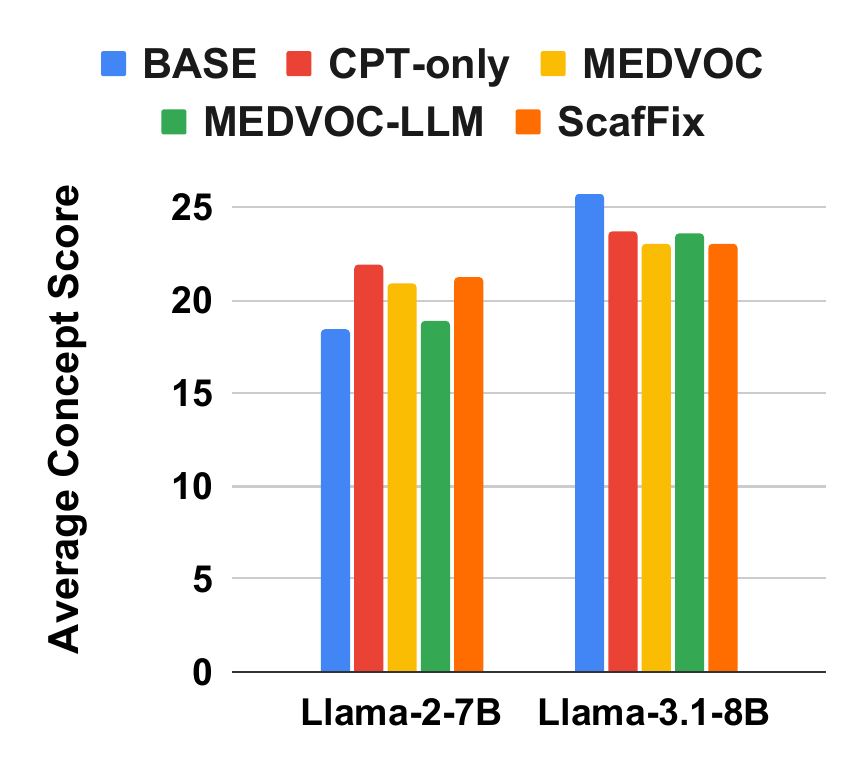}}
    \subfigure[]{\includegraphics[width=0.32\textwidth]{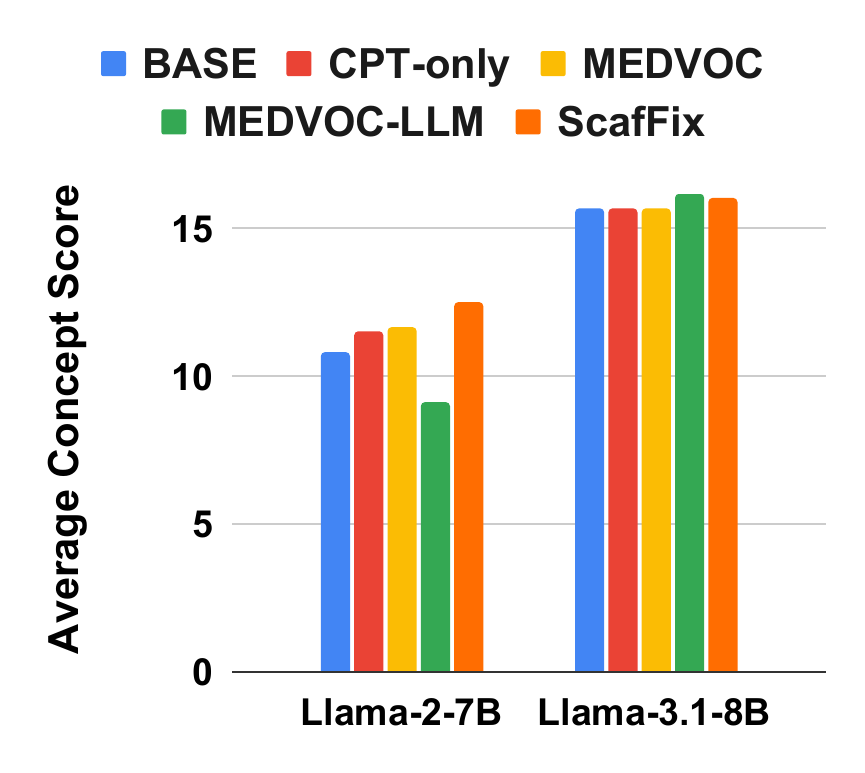}}
    \subfigure[]{\includegraphics[width=0.32\textwidth]{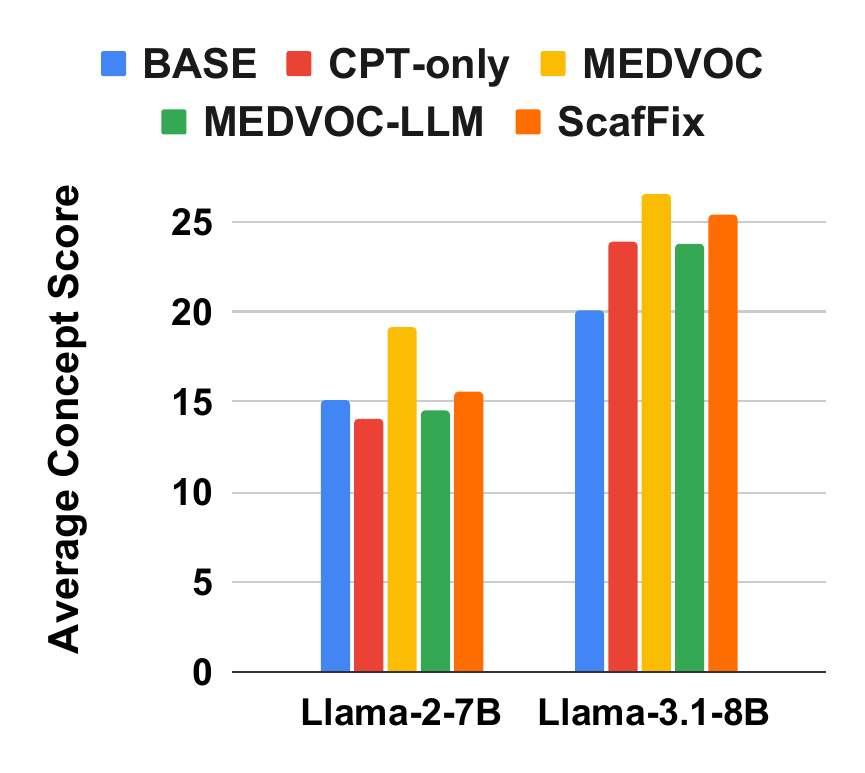}}
    \caption{Concept Score (CSr) observed for \textbf{(a) PubMedQA}; \textbf{(b) EBM}; and \textbf{(c) BioASQ-$\mathcal{M}$} averaged over Difficult\textsubscript{SD}, Difficult\textsubscript{RS}, and Novel\textsubscript{RS} setting. At least one vocabulary adaptation strategy imprves over BASE in a total of 5 out of 6 comparisons. The overall average improvement for $18.75\%$ for Llama-2 and $14.82\%$ for Llama-3.1 models across datasets. }
    \label{fig:results-overall}
\end{figure*}

Table~\ref{tab:results-settings} shows the performance comparison of the different vocabulary adaptation strategies on Llama-2 and Llama-3.1 models. We report the best performance among the two continual pretraining variants of `End-to-End' and `Two-Stage' as previously described in Section~\ref{sec:train-vocab-methods}. The performance values of the individual settings is added to the Appendix~\ref{appendix:results}. Since Mistral vocabulary size is same as Llama-2 (32K), and Qwen-2 vocabulary size is similar as Llama-3.1 (151K and 128K), we add the Mistral (Table~\ref{appendix:tab-mistral-all-cpt-strategies}) and Qwen-2 (Table~\ref{appendix:tab-qwen2-all-cpt-strategies}) results to the Appendix~\ref{appendix:results}. We summarize the key results for Qwen-2 and Mistral model in RQ7. We observe that vocabulary adaptation leads to performance improvement in terms of Rouge-L for Llama-2 and Llama-3.1 in seven out of eight settings (except for Llama-2 on EBM dataset). In terms of Concept-Score which is a proxy measure for faithfulness~\cite{zhang2023famesumm}, Figure~\ref{fig:results-overall} shows that at least one vocabulary adaptation performs the best in five out of six settings, except for Llama-3.1 on PubMedQA dataset.


\subsection{Discussion of Results for Vocabulary Adaptation with LLMs}
We focus on the fine-grained analysis of LLMs in high OOV and high novelty data points. The Rouge-LCS values observed for each such category are reported in Table~\ref{tab:results-settings}. In total, we benchmark an LLM on 48 settings:  32 OOV-related settings and 16 novelty-related settings. Here we report the best performing training method in Difficult\textsubscript{SD}, Difficult\textsubscript{RS} and Novel\textsubscript{RS} setting. The complete results can be found in Table ~\ref{appendix:tab-llama2-all-cpt-strategies} (for Llama-2) and Table~\ref{appendix:tab-llama3-all-cpt-strategies} (for Llama-3.1). The Concept score values are shown in Figure~\ref{fig:results-overall}. We find that at least one vocabulary adaptation strategy improves over BASE in a total of five out of six settings (LLM-Dataset pairs). The overall average improvement for $18.75\%$ for Llama-2 and $14.82\%$ for Llama-3.1 models across datasets. We present two representative examples from EBM and PubMedQA datasets using Llama-2 models in Appendix~\ref{appendix:sec-rep-examples-scaffix}.

Now, we explore research questions to evaluate LLMs in difficult settings and scenarios where vocabulary adaptation strategies does not help.

\paragraph{RQ1: Vocabulary adaptation outperform BASE model on full test data (Test\textsubscript{Full}).} We note that at least one of the vocabulary adaptation strategies (best of MEDVOC-LLM and ScafFix) improves over BASE in five out of eight settings, except for Llama-3.1 PubMedQA dataset. The average performance improvement is $3.68\%$ for Llama-2 and $4.57\%$ for Llama-3.1. The methods where we see improvements are: MEDVOC-LLM (2/5) and ScafFix (3/5) settings. Thus on the entire test set, \textbf{ScafFix}, is the best-performing vocabulary adaptation method. It even outperforms CPT-Only in six out of eight settings by a margin of $2.97\%$.

\paragraph{RQ2: CPT-Only improves over BASE in high novelty and OOV concentration.} We find that CPT-Only (Continual Pretraining without any vocabulary adaptation) improves over BASE in four (Llama-2: 2; Llama-3.1: 2) out of eight high novelty settings. CPT-Only improves over BASE in 11 out of 16 higher OOV settings (Llama-2: 6; Llama-3.1: 5). However, we observe that at least one vocabulary adaptation strategy improves over CPT-Only in high OOV (13 out of 16 settings) and high novelty (in all 8 setting); thus necessitating the need for vocabulary adaptation.

\paragraph{RQ3: Vocabulary adaptation helps in high OOV concentrations in reference summaries and source documents.} We observe that vocabulary adaptation (best of MEDVOC-LLM and ScafFix) outperforms BASE in fourteen out of sixteen settings. For these settings, the average performance improvement over BASE is , $8.74\%$ and $14.64\%$ for Llama-2 and Llama-3.1 respectively. Thus, \textbf{MEDVOC-LLM} is the best vocabulary adaptation strategy in scenarios of high OOV concentration in reference summaries and source documents.

\paragraph{RQ4: Vocabulary adaptation helps in high novelty settings.}  We observe that vocabulary adaptation improves over BASE in six out of eight high novelty settings. The average performance improvement (wherever observed) is $11.92\%$ and $18.03\%$ for Llama-2 and Llama-3.1 when compared to BASE. Thus, \textbf{ScafFix} is the best vocabulary adaptation strategy in high novelty settings as it outperformed in five out of eight settings.

\begin{table}[!ht]
    \centering
    \scriptsize
    \begin{tabular}{p{1.5cm}p{2.5cm}p{2.6cm}}
    \hline
        \textbf{Word} & \textbf{BASE tokenization} & \textbf{ScafFix tokenization} \\ \hline
        microbiologically &  [`Ġmicrobi', `ologically'] & [`\textit{Ġmicro', `biological', `ly'}] \\
        inhibitory & [`Ġinhib', `itory'] & [\textit{`Ġ', `inhibitor', `y'}] \\
        chronically & [`Ġchron', `ically'] & [\textit{`Ġ', `chronic', `ally'}] \\
        antibacterial & [`Ġantib', `acterial'] & [\textit{`Ġanti', `bacteria', `l'}] \\ \hline
    \end{tabular}
    \caption{Samples of words from PubMedQA corpus. For each word, we observe that ScafFix tokenization preserves the morphological boundary while tokenization, unlike BASE tokenization where subwords cross morphological boundaries~\cite{bauwens-delobelle-2024-bpe}.}
    \label{tab:sample_better_tokenization}
\end{table}

\paragraph{RQ5: Comparison of added vocabulary sizes among vocabulary adaptation strategies.} The details of the added vocabulary and fragment score for the different vocabulary adaptation strategies are provided in Table~\ref{tab:vocabulary-stats} (Appendix~\ref{appendix:expt-setup}). We observe that ScafFix reduces fragment score by $30.83\%$ over BASE by adding the minimum amount of added vocabulary terms as compared to the MEDVOC and MEDVOC-LLM. The lowest vocabulary size due to removal of scaffolding tokens leads to a significant drop in the number of under-trained tokens in the LLM vocabulary. This makes the training phase less noisy and yields superior performance during inference.

\paragraph{RQ6: Vocabulary adaptation does not help much in case of extractive summaries, low OOV concentration and low novelty.} We note that all the vocabulary adaptation techniques struggle to outperform both BASE and CPT-Only in PubMedQA (Llama-3.1), BioASQ-$\mathcal{S}$ (Llama-2), and BioASQ-$\mathcal{M}$ (Llama-2 and Llama-3.1). When compared in the entire test set; there is a performance drop of 1.39\% across these two settings. This may be because PubMedQa and BioASQ-$\mathcal{S}$ have low novelty. Specifically, for BioASQ-$\mathcal{S}$, the unigram novelty is just $4.11\%$ (see Table~\ref{tab:dataset-stats}), therefore, the high novelty threshold for top-ten percentile was $35\%$. Thus, we take the average of EBM and PubMedQA high novelty values of $60\%$. This results in vocabulary adaptation outperforming BASE in high novelty setting of BioASQ-$\mathcal{S}$. 

In case of  BioASQ-$\mathcal{M}$, we note that although it is a dataset with higher unigram novelty, the summaries where ScafFix fails have a higher Rouge-L overlap with the Source Document as compared to the summaries where ScafFix does better than BASE, thus making the inference easier irrespective of OOV and Novelty concentration. We provide a detailed error analysis observed in the performance gap in Appendix~\ref{appendix:results} (Table~\ref{appendix:tab-character_eval_BioASQ-M})

\paragraph{RQ7: Proposed vocabulary adaptation methods generalizes to other LLMs such as Qwen-2 and Mistral.} In case of Qwen-2 (Table~\ref{appendix:tab-qwen2-all-cpt-strategies}), we observe that at least one vocabulary adaptation strategies improve over BASE and CPT-Only in 7 out of 12 comparisons. In case of Mistral (Table~\ref{appendix:tab-mistral-all-cpt-strategies}), we observe that at least one of the vocabulary adaptation strategies improved over BASE and CPT-Only in one out of 12 comparisons. In case of both the models (details in Appendix~\ref{appendix:results}), we observe a similar trend with vocabulary adaptation not helping in case of low OOV and novelty concentration (specifically in BioASQ-$\mathcal{S}$) as we observed in RQ6. These findings highlight that our results are potentially generalizable to multiple LLMs. 

\subsection{Human Evaluation}
\begin{figure}[t]
    \centering
\includegraphics[width=0.33\textwidth]{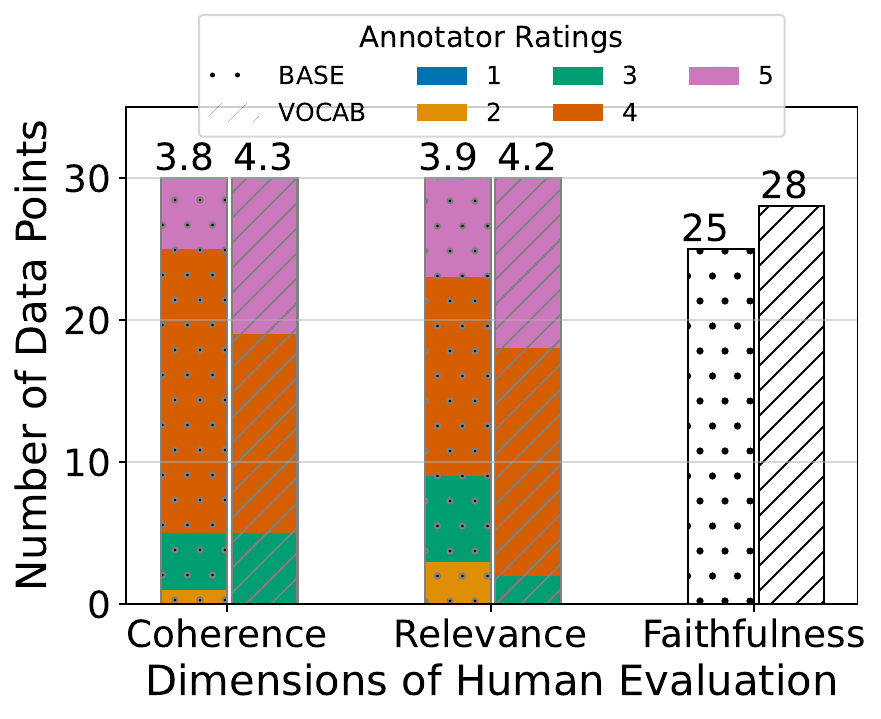}
    \caption{Human evaluation scores in high OOV concentration on `End-to-End' continual pretraining strategy. Vocab corresponds to \textbf{ScafFix} model which produces more relevant, coherent, and faithful summaries during human evaluation with medical experts.}
    \label{fig:human-eval}
\end{figure}

We randomly select $30$ test data points uniformly across the three datasets and two models that have higher expert OOV concentration. We use the Prolific platform to recruit medical experts for annotating summary pairs of ScafFix model and BASE across three aspects~\cite{fabbri-etal-2021-summeval,zhang2023famesumm,balde2024ijcai} namely \textit{relevance}, \textit{coherence} (on a Likert scale of $1$ to $5$), and faithfulness (binary). Each annotator was given $30$ minutes to evaluate $10$ summaries and was compensated at a rate of $8$ UK pounds per hour (see Appendix~\ref{appendix:human-eval-app} for more details), and each summary par was evaluated by three annotators. Figure~\ref{fig:human-eval} shows the human evaluation results where the ScafFix method generates more faithful summaries ($93.34\%$ versus $83.34\%$ of summaries are faithful), and more relevant summaries, where $93.34\%$ of data points get a score $\ge 4$ in Likert scale, as compared to $70\%$ by BASE.

\section{Related Works}
\paragraph{Vocabulary Expansion in LLMs.} Recent research has focused on enhancing large language models (LLMs) through domain-specific vocabulary optimization. ~\citet{liu-etal-2024-gold} introduces VEGAD, an adaptive method for selecting an optimal subset of domain vocabulary, which enhances performance on both specialized and general tasks, validated on Chinese datasets. Similarly, ~\citet{gao2024ve} introduced VE-KD, a method combining vocabulary expansion and knowledge distillation to train efficient domain-specific language models for BioBERT and PubMedBERT on biomedical tasks. These approaches align with broader efforts, such as, Chatlaw~\cite{cui2024chatlawmultiagentcollaborativelegal}, an AI legal assistant, that employed a knowledge graph-enhanced mixture-of-experts model to address legal domain challenges, and ~\cite{chinesellama}, who developed efficient text encoding strategies for Chinese LLMs. Similar efforts have been put in other languages~\cite{nag-etal-2024-cost,tejaswi-etal-2024-exploring,yamaguchi2024can}. However, none of the works benchmarked the effect in the fine-grained manner that we aimed to show in this study.

\paragraph{Over-fragmentation in Medical domain.} Over-fragmentation (spliting of domain words in more than one subword) in tokenization is a significant challenge in adapting large language models (LLMs) to the medical domain due to specialized terminology. ~\citet{si2019enhancing} highlighted the limitations of traditional tokenization on the task of clinical concept extraction where they show models like BERT poorly toknize medical named entities. ~\citet{nguyen2019investigating} emphasized the impact of lexical segmentation on transformer-based models and suggested that not only using a domin-specific vocabulary helps but also a continual pretraining phase helps. ~\citet{yuan-etal-2022-biobart} introduced BioBART and ~\citet{labrak-etal-2024-biomistral} introduced BioMistral, biomedical-specific generative model designed to address domain-specific tokenization challenges that outperformed various SoTA methods on different kinds of task like summarization and question-answering. ~\citet{liu-etal-2023-task} proposed task-adaptive tokenization to enhance long-form text generation by dynamically adjusting tokenization for domain-specific semantics by going beyond word boundaries during tokenization. Our work builds on these advancements by benchmarking the effect of refining tokenization on recent Llama-2, Llama-3, Mistral, and Qwen-2 models.
\section{Conclusion}
This work is a first step towards understanding how vocabulary adaptation strategies effect the performance of general-purpose LLMs in medical text summarization. We first show that general-purpose LLMs fail in certain challenging generation scenarios where reference summaries have high OOV concentration and high novelty. We then benchmark the performance of three vocabulary adaptation strategies on four models: Llama-2 7B, Mistral 7B, Qwen-2 7B, and Llama-3.1 8B model; over three biomedical summarization datasets and two continual pretraining strategies. Llama-3.1 and Qwen-2 even with a vocabulary size of around 128K and 151K tokens, still faces the over-fragmentation issue with medical words, and vocabulary adaptation is shown to help improve the LLM summarization performance. Medical experts find that vocabulary adaptation improves the relevance, coherence and faithfulness of medical summaries. We make the codebase available for reproducibility purposes.



\section{Limitations}
We identify two main limitations of this work
. Firstly, because of hardware constraints we could not explore much larger variants (like 13B and 70B) of the models considered in the study. Second, we note that all the results provided in this paper are generated by applying LoRA during the pretraining phase. This might make a  difference when replicating the results with full-scale fine-tuning without any parameter efficient fine-tuning strategies. 

\section{Ethics Statement and Broader Impact}
It is a well known fact that LLMs are often prone to hallucinations, producing outputs that could not be verified as-is from the source context without any domain-specific knowledge. While the proposed methods here does improve faithfulness, the quality of the summaries generated is still not ready for deployment to the public without evaluating the safety perspective of such responses. We believe more research is needed to align the outputs of these LLMs for such high-stake domains like healthcare where a single misinformation (or disinformation) could lead to drastic consequences.

\section*{Acknowledgments}
We thank the Ministry of Education, Govt of India, for supporting Gunjan Balde with Prime Minister Research Fellowship during his Ph.D. tenure. This research was partially funded by a Google Academic Research Award.
\bibliography{custom}

\appendix
\section{Experimental Setup}\label{appendix:expt-setup}

\paragraph{Vocabulary Sizes.} We report the vocabulary sizes along with the resultant fragment score in Table~\ref{tab:vocabulary-stats}.

\begin{table}[!ht]
\centering
\setlength{\tabcolsep}{0.1cm}
\scalebox{0.6}{
\begin{tabular}{p{2cm}rr|rr|rr|rr}
\hline
 \multicolumn{1}{c}{\textbf{Model}}  & \multicolumn{2}{c|}{\textbf{Llama-2}} & \multicolumn{2}{c|}{\textbf{Mistral}} & \multicolumn{2}{c|}{\textbf{Llama-3.1}} & \multicolumn{2}{c}{\textbf{Qwen-2}} \\
& \multicolumn{1}{c}{\textbf{|V|}} & \multicolumn{1}{c|}{\textbf{FragSr}} & \multicolumn{1}{c}{\textbf{|V|}} & \multicolumn{1}{c|}{\textbf{FragSr}} & \multicolumn{1}{c}{\textbf{|V|}} & \multicolumn{1}{c|}{\textbf{FragSr}} & \multicolumn{1}{c}{\textbf{|V|}} & \multicolumn{1}{c}{\textbf{FragSr}} \\ \hline
\multicolumn{9}{c}{\textbf{PubMedQA}} \\
CPT-Only & 32000 & 2.53 & 32000 & 2.58 & 128256 & 2.29 & 151646 & 2.29 \\
MEDVOC & 38245 & 1.48 & 37642 & 1.55 & 129078 & 1.21 & 152441 & 1.21 \\
MEDVOC-LLM & 34572 & 1.66 & 34219 & 1.73 & 128808 & 1.25 & 152131 & 1.25 \\
ScafFix & 32200 & 2.14 & 32266 & 2.20 & 128456 & 1.31 & 152000 & 1.28 \\ \hline

\multicolumn{9}{c}{\textbf{EBM}} \\
CPT-Only & 32000 & 2.65 & 32000 & 2.75 & 128256 & 2.33 & 151646 & 2.33 \\
MEDVOC & 42836 & 1.48 & 43908 & 1.61 & 131231 & 1.12 & 154092 & 1.13 \\
MEDVOC-LLM & 34572 & 1.66 & 36252 & 1.80 & 130253 & 1.22 & 153468  & 1.23 \\
ScafFix & 32150 & 1.83 & 32060 & 2.54 & 128456 & 1.37 & 152062 & 1.35 \\ \hline

\multicolumn{9}{c}{\textbf{BioASQ}} \\
CPT-Only & 32000 & 2.98 & 32000 & 2.69 & 128256 & 2.44 & 151646 & 2.65 \\
MEDVOC & 43194 & 1.93 & 46502 & 1.44 & 133099 & 1.24 & 155518 & 1.26 \\
MEDVOC-LLM & 37399 & 2.18 & 37506 & 1.65 & 130966 & 1.30 & 154612 & 1.30 \\
ScafFix & 32300 & 2.43 & 32193 & 2.39 & 128506 & 1.56 & 152098 & 1.44 \\ \hline
\end{tabular}}
\caption{The final vocabulary sizes (|V|) along with the resultant fragment score (\textbf{FragSr}) observed on Medical OOV words. MEDVOC, MEDVOC-{LLM}, and ScafFix are the best vocabulary adaptation scheme which has the least vocabulary sizes and decent fragment score.}
\label{tab:vocabulary-stats}
\end{table}

\paragraph{Prompts used.} We use the prompt template inspired from ClinSumm~\cite{van2024adapted} and \citet{zhang2025beyond} as shown in Table~\ref{appendix:tab-prompt}. 

\begin{table}[t]
    \centering
    \scriptsize
    \scalebox{0.95}{
    \begin{tabular}{p{7cm}}
        \hline
        \multicolumn{1}{c}{\textbf{Prompt}}  \\
        \hline
        You are a medical expert. You are given a query and query-relevant information as inputs. Your task is to summarize this information. The summary should be concise, include only non-redundant, query-relevant evidence, and be approximately 100 words long. Use the provided examples to guide word choice. \\

        \multicolumn{1}{c}{\textit{--n icl examples concatenated using `\#\#'--}} \\
        Query \{i\}: \{Train-Query\} \\
        Document \{i\}: \{Train-Source Document\}  \\
        Summary \{i\} : \{Train-Summary\} \\
        \#\# \\

        \multicolumn{1}{c}{\textit{--Test Example--}} \\
        Query: \{Test-Query\} \\
        Document: \{Test-Source Document\} \\
        Summary: \\ \hline
    \end{tabular}}
    \caption{Prompt Template used to prompt the language models for the task of query focused summarization.}
    \label{appendix:tab-prompt}
\end{table}

\section{Results} \label{appendix:results}

\begin{table}[!ht]
    \centering
    \scriptsize
    \scalebox{0.95}{
    \begin{tabular}{p{5cm}|l|l} \hline
        \textbf{Metric} & \textbf{Subset-1} & \textbf{Subset-2}  \\ \hline
        \multicolumn{3}{c}{\textbf{Llama-2}} \\
        Difficult-RS Concentration & 9.78\% & 10.77\% \\
        Novel-RS Concentration & 40.65\% & 41.92\% \\
        Rouge-LCS overlap between source and reference & 38.17 &35.65 \\ \hline

        \multicolumn{3}{c}{\textbf{Llama-3.1}} \\
        Difficult-RS Concentration & 7.47\%	 & 8.81\% \\
        Novel-RS Concentration & 39.46\%	& 42.78\% \\
        Rouge-LCS overlap between source and reference &	39.52 & 34.27 \\ \hline
        
    \end{tabular}}
    \caption{ Difference of characteristics between instances where BASE has better Rouge-LCS than ScafFix (Subset-1) and instances where ScafFix has higher Rouge-LCS than BASE (Subset-2). The instances where BASE has better Rouge-LCS than ScafFix (Subset-1) have lesser values of Difficult-RS Concentration as well as Novel-RS Concentration, but higher values of Rouge-LCS overlap compared to Subset-2.}
    \label{appendix:tab-character_eval_BioASQ-M}
\end{table}

\begin{table*}[!ht]
\centering
\scalebox{0.6}{
\begin{tabular}{lrrrrrr|rrrrrr}
\hline
 & \multicolumn{6}{c|}{\textbf{End-To-End}} & \multicolumn{6}{c}{\textbf{Two-Stage}} \\
 & \multicolumn{1}{l}{\textbf{Test\textsubscript{Full}}} & \multicolumn{1}{l}{\textbf{Difficult\textsubscript{SD}}} & \multicolumn{1}{l}{\textbf{Difficult\textsubscript{RS}}} & \multicolumn{1}{l}{\textbf{Novel\textsubscript{RS}}} & \multicolumn{1}{l}{\textbf{All\textsubscript{SD}}} & \multicolumn{1}{l|}{\textbf{All\textsubscript{RS}}} & \multicolumn{1}{l}{\textbf{Test\textsubscript{Full}}} & \multicolumn{1}{l}{\textbf{Difficult\textsubscript{SD}}} & \multicolumn{1}{l}{\textbf{Difficult\textsubscript{RS}}} & \multicolumn{1}{l}{\textbf{Novel\textsubscript{RS}}} & \multicolumn{1}{l}{\textbf{All\textsubscript{SD}}} & \multicolumn{1}{l}{\textbf{All\textsubscript{RS}}} \\ \hline

\multicolumn{13}{c}{\textbf{PubMedQA}} \\ 
BASE & 26.33 & 23.53 & 24.40 & 19.78 & 25.46 & 27.45 & 26.33 & 23.53 & 24.40 & 19.78 & 22.64 & 27.45 \\
CPT-only & 27.12 & 25.97 & 28.00 & 22.21 & \textbf{27.86} & 27.59 & 27.11 & 25.00 & 26.92 & 21.24 & \textbf{24.07} & 26.67 \\
MEDVOC & 26.50 & 24.35 & 22.88 & 20.14 & 24.59 & 27.70 & 26.39 & 24.00 & 24.82 & 20.88 & 22.22 & 26.23 \\
MEDVOC-LLM & 26.90 & 24.49 & 26.67 & 20.00 & 25.27 & 27.12 & \textbf{27.30} & \textbf{25.54} & \textbf{27.45} & 20.84 & 21.43 & 28.89 \\
ScafFix & \textbf{27.61} & \textbf{26.32} & \textbf{28.57} & \textbf{22.22} & 25.83 & \textbf{28.57} & 27.05 & 25.00 & 24.75 & \textbf{21.30} & 21.28 & \textbf{30.19} \\ \hline
\multicolumn{13}{c}{\textbf{EBM}} \\
BASE & 18.56 & 17.33 & \textbf{16.00} & 11.20 & 14.95 & \textbf{15.69} & 18.56 & 17.33 & \textbf{16.00} & 11.20 & 14.95 & \textbf{15.69} \\
CPT-only & \textbf{18.92} & \textbf{18.18} & 15.19 & 11.77 & 14.64 & 14.04 & 19.13 & \textbf{19.05} & 15.87 & 11.77 & 14.29 & 13.95 \\
MEDVOC & 18.09 & \textbf{18.18} & 14.63 & 11.91 & \textbf{15.31} & 14.74 & 18.60 & 18.65 & 15.38 & 12.50 & 15.83 & 14.29 \\
MEDVOC-LLM & 18.77 & 17.20 & 14.55 & 12.50 & 14.45 & 15.27 & \textbf{19.27} & 17.54 & 14.71 & \textbf{12.90} & \textbf{17.55} & 14.81 \\
ScafFix & 18.67 & 16.67 & 15.39 & \textbf{13.80} & 15.27 & 15.63 & 18.65 & 17.40 & 14.71 & 11.77 & 15.47 & 14.71 \\ \hline
\multicolumn{13}{c}{\textbf{BioASQ-M}} \\
BASE & \textbf{28.50} & 28.00 & 27.27 & \textbf{21.53} & \textbf{26.71} & 27.27 & \textbf{28.50} & 28.00 & 27.27 & \textbf{21.53} & \textbf{26.71} & 27.27 \\
CPT-only & 27.22 & 28.57 & 27.91 & 20.31 & 24.44 & 27.91 & 26.73 & \textbf{29.41} & 27.91 & 19.80 & 25.20 & \textbf{28.57} \\
MEDVOC & 24.19 & 26.32 & 23.26 & 16.53 & 23.08 & 26.09 & 24.47 & 28.07 & 21.74 & 16.06 & 24.19 & 24.57 \\
MEDVOC-LLM & 24.50 & 28.00 & 25.00 & 18.90 & 23.76 & 26.38 & 25.15 & 27.59 & 25.00 & 16.85 & 23.17 & 24.78 \\
ScafFix & 26.16 & \textbf{30.00} & \textbf{29.79} & 21.17 & 22.84 & \textbf{28.07} & 26.00 & 28.57 & \textbf{28.57} & 19.53 & 24.10 & 26.49 \\ \hline
\multicolumn{13}{c}{\textbf{BioASQ-S}} \\
BASE & 32.12 & 35.72 & 26.32 & \textbf{20.59} & \textbf{33.33} & \textbf{27.92} & 32.12 & 35.72 & 26.12 & 20.53 & \textbf{33.33} & \textbf{27.92} \\
CPT-only & 33.09 & 29.35 & 25.53 & 19.64 & 23.65 & 24.12 & \textbf{33.30} & 30.33 & 29.22 & 20.00 & 30.30 & 21.92 \\
MEDVOC & 32.34 & 30.26 & 29.41 & 20.29 & 29.41 & 22.47 & 32.26 & 27.40 & 28.17 & 18.18 & 26.09 & 23.53 \\
MEDVOC-LLM & 32.75 & 34.52 & 30.30 & 29.29 & 33.33 & 24.47 & 32.40 & 30.33 & \textbf{29.41} & 20.29 & 27.03 & 21.53 \\
ScafFix & \textbf{33.34} & \textbf{36.44} & \textbf{31.25} & 19.05 & 27.03 & 24.00 & 32.88 & \textbf{36.44} & 29.29 & \textbf{22.22} & \textbf{33.33} & 24.00 \\ \hline
\end{tabular}}
\caption{Performance comparison in terms of Rouge-L (R-L) between the two continual pretraining strategies of `End-to-End' and `Two-Stage' on Llama-2 7B model.}
\label{appendix:tab-llama2-all-cpt-strategies}
\end{table*}

\begin{table*}[t]
\centering
\scalebox{0.6}{
\begin{tabular}{lrrrrrr|rrrrrr}
\hline
 & \multicolumn{6}{c|}{\textbf{End-To-End}} & \multicolumn{6}{c}{\textbf{Two-Stage}} \\
 & \multicolumn{1}{l}{\textbf{Test\textsubscript{Full}}} & \multicolumn{1}{l}{\textbf{Difficult\textsubscript{SD}}} & \multicolumn{1}{l}{\textbf{Difficult\textsubscript{RS}}} & \multicolumn{1}{l}{\textbf{Novel\textsubscript{RS}}} & \multicolumn{1}{l}{\textbf{All\textsubscript{SD}}} & \multicolumn{1}{l|}{\textbf{All\textsubscript{RS}}} & \multicolumn{1}{l}{\textbf{Test\textsubscript{Full}}} & \multicolumn{1}{l}{\textbf{Difficult\textsubscript{SD}}} & \multicolumn{1}{l}{\textbf{Difficult\textsubscript{RS}}} & \multicolumn{1}{l}{\textbf{Novel\textsubscript{RS}}} & \multicolumn{1}{l}{\textbf{All\textsubscript{SD}}} & \multicolumn{1}{l}{\textbf{All\textsubscript{RS}}} \\ \hline
\multicolumn{13}{c}{\textbf{PubMedQA}} \\
BASE & \textbf{28.10} & 24.46 & 26.87 & 18.87 & \textbf{28.89} & 29.21 & \textbf{28.10} & 24.46 & 26.87 & 18.87 & 29.23 & 32.10 \\
CPT-only & 26.62 & 24.52 & \textbf{27.08} & 21.07 & 25.93 & 26.74 & 27.05 & \textbf{27.71} & 26.97 & 19.64 & \textbf{26.71} & 28.65 \\
MEDVOC & 27.86 & 24.68 & 26.23 & 21.87 & 25.97 & \textbf{31.12} & 27.17 & 25.76 & \textbf{27.12} & 19.25 & 28.06 & 29.62 \\
MEDVOC-LLM & 27.69 & \textbf{26.51} & 26.67 & 21.74 & 25.53 & 30.43 & 27.15 & 25.19 & 26.51 & 19.86 & 28.45 & 28.51 \\
ScafFix & 27.67 & 23.85 & 25.00 & \textbf{22.22} & 25.00 & 29.16 & 27.25 & 26.37 & 26.42 & \textbf{21.16} & 28.16 & \textbf{28.41} \\ \hline

\multicolumn{13}{c}{\textbf{EBM}} \\
BASE & 20.04 & 15.79 & 14.54 & 11.37 & 15.62 & 15.18 & 20.04 & 15.79 & 14.54 & 11.37 & 15.62 & 15.18 \\
CPT-only & 20.13 & 16.98 & 17.15 & 13.33 & 13.79 & 16.67 & 20.20 & 17.24 & 14.94 & 11.90 & 14.81 & 13.82 \\
MEDVOC & 20.31 & 17.93 & 16.03 & 12.17 & 13.56 & 16.87 & 20.45 & \textbf{17.65} & 15.56 & 13.34 & \textbf{16.36} & \textbf{15.78} \\
MEDVOC-LLM & 20.75 & \textbf{18.39} & 16.40 & 13.04 & \textbf{16.13} & \textbf{17.05} & 20.42 & \textbf{17.65} & 15.63 & \textbf{14.82} & 15.79 & 15.38 \\
ScafFix & \textbf{20.79} & 16.18 & \textbf{17.17} & \textbf{13.80} & 15.39 & 14.29 & \textbf{20.50} & 17.07 & \textbf{16.85} & 13.04 & 16.13 & 14.76 \\ \hline

\multicolumn{13}{c}{\textbf{BioASQ-M}} \\
BASE & \textbf{29.28} & 28.57 & 26.67 & \textbf{21.51} & 25.54 & 23.77 & \textbf{29.28} & 28.57 & 26.67 & \textbf{21.51} & \textbf{25.53} & 23.77 \\
CPT-only & 27.56 & 28.57 & 26.09 & 19.84 & 25.54 & 23.53 & 27.25 & 28.57 & \textbf{29.79} & 19.14 & 24.56 & \textbf{27.73} \\
MEDVOC & 27.71 & 28.57 & 27.27 & 20.46 & 24.39 & 25.86 & 27.18 & 30.00 & 26.93 & 19.33 & 24.56 & 26.84 \\
MEDVOC-LLM & 27.45 & 29.17 & 27.27 & 19.05 & 25.00 & 26.09 & 27.92 & \textbf{30.43} & 29.03 & 20.97 & 25.00 & 27.36 \\
ScafFix & 28.91 & \textbf{30.00} & \textbf{27.59} & 21.23 & \textbf{25.89} & \textbf{26.45} & 27.86 & 28.13 & 28.57 & 20.94 & 24.10 & 27.38 \\ \hline

\multicolumn{13}{c}{\textbf{BioASQ-S}} \\
BASE & 35.25 & 27.03 & 29.60 & 18.18 & 30.15 & \textbf{27.92} & 35.25 & 27.03 & 29.60 & 18.18 & 30.15 & 27.92 \\
CPT-only & 37.60 & \textbf{42.42} & 30.04 & 18.67 & \textbf{42.02} & 31.58 & 36.01 & 29.41 & 24.70 & 17.14 & 31.79 & 28.57 \\
MEDVOC & 37.00 & 31.25 & 27.03 & 20.29 & 30.38 & 28.92 & 37.01 & 32.43 & 29.79 & 18.18 & 32.43 & 31.11 \\
MEDVOC-LLM & 27.43 & 35.09 & 29.34 & 20.29 & 33.00 & 30.37 & \textbf{37.15} & \textbf{37.84} & \textbf{32.89} & 20.29 & 31.01 & 35.04 \\
ScafFix & \textbf{37.22} & 37.04 & \textbf{33.34} & \textbf{20.34} & 35.91 & \textbf{35.02} & 36.70 & 34.15 & 32.26 & \textbf{20.90} & \textbf{33.74} & \textbf{33.35} \\ \hline
\end{tabular}}
\caption{Performance comparison in terms of Rouge-L (R-L) between the two continual pretraining strategies of `End-to-End' and `Two-Stage' on Llama-3.1 8B model}
\label{appendix:tab-llama3-all-cpt-strategies}
\end{table*}

\begin{table*}[t]
\centering
\scalebox{0.6}{
\begin{tabular}{lrrrrrr|rrrrrr}
\hline
 & \multicolumn{6}{c|}{\textbf{End-To-End}} & \multicolumn{6}{c}{\textbf{Two-Stage}} \\
 & \multicolumn{1}{l}{\textbf{Test\textsubscript{Full}}} & \multicolumn{1}{l}{\textbf{Difficult\textsubscript{SD}}} & \multicolumn{1}{l}{\textbf{Difficult\textsubscript{RS}}} & \multicolumn{1}{l}{\textbf{Novel\textsubscript{RS}}} & \multicolumn{1}{l}{\textbf{All\textsubscript{SD}}} & \multicolumn{1}{l|}{\textbf{All\textsubscript{RS}}} & \multicolumn{1}{l}{\textbf{Test\textsubscript{Full}}} & \multicolumn{1}{l}{\textbf{Difficult\textsubscript{SD}}} & \multicolumn{1}{l}{\textbf{Difficult\textsubscript{RS}}} & \multicolumn{1}{l}{\textbf{Novel\textsubscript{RS}}} & \multicolumn{1}{l}{\textbf{All\textsubscript{SD}}} & \multicolumn{1}{l}{\textbf{All\textsubscript{RS}}} \\ \hline

\multicolumn{13}{c}{\textbf{PubMedQA}} \\ 
BASE       & 25.40 & \textbf{27.04} & \textbf{28.32} & \textbf{19.48} & 22.97 & 29.29 & 25.40 & \textbf{27.04} & \textbf{28.32} & 19.48 & 22.97 & \textbf{29.29} \\
CPT-Only   & 25.00 & 25.00 & 25.23 & 19.14 & 24.64 & 27.80 & \textbf{25.51} & 25.28 & 26.67 & \textbf{19.84} & \textbf{24.81} & 28.37 \\
MEDVOC     & 24.49 & 24.09 & 26.67 & 17.56 & 23.88 & 28.27 & 24.64 & 24.99 & 25.01 & 18.85 & 22.79 & 26.49 \\
MEDVOC-LLM & \textbf{25.45} & 25.05 & 24.58 & 19.00 & 24.19 & \textbf{29.62} & 24.47 & 24.06 & 24.25 & 19.20 & 23.49 & 28.10 \\
ScafFix    & 25.40 & 24.04 & 25.27 & 18.67 & \textbf{24.62} & 27.92 & 24.56 & 24.78 & 24.85 & 17.79 & 23.14 & 28.29 \\ \hline

\multicolumn{13}{c}{\textbf{EBM}} \\
BASE       & \textbf{17.43} & \textbf{17.86} &\textbf{17.06} & 11.49 & \textbf{16.87} & 12.33 & \textbf{17.43} & \textbf{17.86} & \textbf{17.06} & 11.49 & \textbf{16.87} & 13.33 \\
CPT-Only   & 17.15 & 17.70 & 16.49 & \textbf{12.12} & 15.85 & \textbf{13.79} & 17.39 & 17.82 & 15.59 & 11.32 & 14.29 & 12.77 \\
MEDVOC     & 17.63 & 16.09 & 14.04 & 11.24 & 15.46 & 12.99 & 17.10 & 15.83 & 14.61 & \textbf{13.01} & 15.38 & \textbf{13.70} \\
MEDVOC-LLM & 17.19 & 17.40 & 14.77 & 11.11 & 15.19 & 13.33 & 16.85 & 16.22 & 15.48 & 12.66 & 15.19 & 12.50 \\
ScafFix    & 16.80 & 17.19 & 14.98 & 11.19 & 15.22 & 13.11 & 17.21 & 16.87 & 13.30 & 11.99 & 13.33 & 12.90 \\ \hline

\multicolumn{13}{c}{\textbf{BioASQ-M}} \\
BASE       & \textbf{25.97} & \textbf{28.21} & \textbf{25.64} & \textbf{21.88} & \textbf{26.23} & \textbf{26.67} & \textbf{25.97} & \textbf{28.21} & \textbf{25.64} & \textbf{21.88} & \textbf{26.23} & \textbf{26.67} \\ 
CPT-Only   & 22.73 & 25.64 & 25.00 & 21.05 & 23.38 & 23.53 & 23.53 & 27.03 & 24.56 & 21.54 & 23.26 & 24.56 \\    
MEDVOC     & 21.54 & 23.68 & 17.02 & 16.67 & 22.20 & 21.43 & 21.95 & 25.30 & 15.38 & 17.02 & 22.95 & 22.22 \\ 
MEDVOC-LLM & 22.73 & 24.69 & 20.00 & 16.18 & 23.08 & 22.73 & 22.64 & 26.67 & 19.23 & 17.91 & 24.00 & 22.86 \\ 
ScafFix    & 22.73 & 24.39 & 19.35 & 18.52 & 22.58 & 23.81 & 20.69 & 23.53 & 17.02 & 16.33 & 22.22 & 21.28 \\ \hline

\multicolumn{13}{c}{\textbf{BioASQ-S}} \\
BASE       & \textbf{35.44} & \textbf{53.59} & \textbf{40.00} & \textbf{28.57} & \textbf{40.00} & \textbf{40.00} & \textbf{35.44} & \textbf{53.59} & \textbf{40.00} & \textbf{28.57} & \textbf{40.00} & \textbf{40.00} \\
CPT-Only   & 28.57 & 31.01 & 23.58 & 22.22 & 28.57 & 25.81 & 29.63 & 34.14 & 28.57 & 21.15 & 32.26 & 28.57 \\ 
MEDVOC     & 28.57 & 31.16 & 26.32 & 19.15 & 27.03 & 27.78 & 27.78 & 29.61 & 25.93 & 19.61 & 28.00 & 23.81 \\
MEDVOC-LLM & 28.17 & 35.19 & 27.91 & 20.22 & 35.90 & 29.63 & 27.27 & 34.22 & 26.09 & 20.51 & 30.30 & 26.67 \\
ScafFix    & 30.00 & 33.52 & 29.63 & 25.00 & 32.00 & 33.33 & 24.69 & 31.95 & 24.24 & 20.00 & 22.22 & 25.00 \\ \hline
\end{tabular}}
\caption{Performance comparison in terms of Rouge-L (R-L) between the two continual pretraining strategies of `End-to-End' and `Two-Stage' on Mistral 7B model.}
\label{appendix:tab-mistral-all-cpt-strategies}
\end{table*}

\begin{table*}[t]
\centering
\scalebox{0.6}{
\begin{tabular}{lrrrrrr|rrrrrr}
\hline
 & \multicolumn{6}{c|}{\textbf{End-To-End}} & \multicolumn{6}{c}{\textbf{Two-Stage}} \\
 & \multicolumn{1}{l}{\textbf{Test\textsubscript{Full}}} & \multicolumn{1}{l}{\textbf{Difficult\textsubscript{SD}}} & \multicolumn{1}{l}{\textbf{Difficult\textsubscript{RS}}} & \multicolumn{1}{l}{\textbf{Novel\textsubscript{RS}}} & \multicolumn{1}{l}{\textbf{All\textsubscript{SD}}} & \multicolumn{1}{l|}{\textbf{All\textsubscript{RS}}} & \multicolumn{1}{l}{\textbf{Test\textsubscript{Full}}} & \multicolumn{1}{l}{\textbf{Difficult\textsubscript{SD}}} & \multicolumn{1}{l}{\textbf{Difficult\textsubscript{RS}}} & \multicolumn{1}{l}{\textbf{Novel\textsubscript{RS}}} & \multicolumn{1}{l}{\textbf{All\textsubscript{SD}}} & \multicolumn{1}{l}{\textbf{All\textsubscript{RS}}} \\ \hline

\multicolumn{13}{c}{\textbf{PubMedQA}} \\ 
BASE & 24.00 & 22.22 & 24.14 & 18.82 & 23.74 & \textbf{30.00} & 24.00 & 22.22 & 24.14 & 18.82 & 23.74 & \textbf{30.00} \\
CPT-only & 24.65 & \textbf{27.85} & 25.39 & 17.50 & \textbf{24.85} & 25.81 & 24.73 & 23.88 & 24.78 & 17.65 & \textbf{25.41} & 25.64 \\
MEDVOC & 24.59 & 23.08 & \textbf{25.93} & 18.37 & 23.21 & 26.32 & 24.89 & 25.45 & 26.20 & 18.18 & 23.57 & 25.00 \\
MEDVOC-LLM & \textbf{25.00} & 25.00 & 25.90 & \textbf{19.50} & 24.83 & 26.67 & \textbf{25.00} &\textbf{25.00} & \textbf{27.06} & 16.87 & 23.47 & 27.60 \\
ScafFix &  24.66 & 25.58 & 25.65 & 18.60 & 23.81 & 25.54 & \textbf{25.00} & \textbf{25.00} & 26.32 & \textbf{18.87} & 25.00 &  26.32 \\ \hline

\multicolumn{13}{c}{\textbf{EBM}} \\
BASE & 17.52 & 16.49 & 14.04 & 11.43 & 16.07 & 14.46 & 17.52 & 16.49 & 14.04 & 11.43 & 16.07 & 14.46\\
CPT-only & 17.60 & 15.38 & 15.17 & 11.86 & 16.42 & 13.91 & \textbf{17.80} & 14.93 & 14.89 & 11.51 & 14.77 & 13.95 \\
MEDVOC & 17.54 & 16.13 & 16.13 & 12.31 & 16.67 & 15.62 & 17.43 & 15.87 & 15.47 & \textbf{12.50} & \textbf{16.75} & 15.15\\
MEDVOC-LLM & 17.34 & 16.13 & 14.70 & 12.60 & 16.16 & 15.15 & 17.65 & \textbf{17.93} & 15.39 & 11.76 & 16.39 & 15.15 \\
ScafFix & \textbf{17.78} & \textbf{17.24} & \textbf{17.64} & \textbf{13.33} & \textbf{16.98} & \textbf{17.58} & 17.67 & 17.24 & \textbf{16.67} & 12.04 & 16.07 & \textbf{17.14} \\ \hline

\multicolumn{13}{c}{\textbf{BioASQ-$\mathcal{M}$}} \\
BASE & \textbf{25.45} & 30.00 & \textbf{27.27} & 19.67 & 26.09 & \textbf{27.73} & \textbf{25.45} & \textbf{30.00} & \textbf{27.27} & 19.67 & \textbf{26.09} & \textbf{27.73} \\
CPT-only & 23.88 & 28.57 & 26.10 & 18.52 & 25.25 & 26.90 & 24.56 & 28.57 & 25.00 & 17.86 & 25.35 & 26.90 \\
MEDVOC & 25.00 & \textbf{30.20} & 26.67 & \textbf{20.00} & \textbf{26.67} & 26.85 & 24.49 & 28.57 & 25.00 & 19.05 & 24.70 & 26.32  \\
MEDVOC-LLM & 24.30 & 28.57 & 25.71 & 19.23 & 25.35 & 27.27 & 24.34 & 28.57 & 26.67 & \textbf{20.00} & 25.00 & 26.80 \\
ScafFix & 24.49 & 28.24 & 26.23 & 19.67 & 25.00 & 27.20 & 24.69 & 28.57 & 26.67 & \textbf{20.00} & 25.00 & 27.27 \\ \hline

\multicolumn{13}{c}{\textbf{BioASQ-$\mathcal{S}$}} \\
BASE & \textbf{36.19} & \textbf{45.31} & \textbf{45.24} & \textbf{28.57} & \textbf{41.89} & \textbf{43.17} & \textbf{36.19} & \textbf{45.31} & \textbf{45.24} & \textbf{28.57} & \textbf{41.89} & \textbf{43.17}\\
CPT-only & 32.43 & 37.89 & 28.53 & 24.49 & 35.59& 29.18 & 31.59 & 31.50 & 32.57 & 25.00 & 34.46& 29.90 \\
MEDVOC & 31.91 & 35.56 & 32.46 & 24.62 & 32.95 & 31.29 & 31.59 & 33.69 & 31.86 & 25.00 & 33.81 & 30.25 \\
MEDVOC-LLM & 32.14 & 30.49 & 30.94 & 26.32 & 33.81 & 30.77 & 32.14 & 37.97 & 31.84 & 24.56 & 36.18 & 34.29  \\
ScafFix & 32.00 & 39.36 & 36.94 & 26.67 & 37.23 & 31.01 & 31.58 & 38.28 & 32.13 & 26.00 & 36.04 & 29.10 \\ \hline
\end{tabular}}
\caption{Performance comparison in terms of Rouge-L (R-L) between the two continual pretraining strategies of `End-to-End' and `Two-Stage' on Qwen2-7B model.}
\label{appendix:tab-qwen2-all-cpt-strategies}
\end{table*}

\paragraph{Comparing Continual Pretraining Strategies for Llama-2 and Llama-3.1.}In Table~\ref{appendix:tab-llama2-all-cpt-strategies}, we report the performance we observe using two different continual pretraining strategies on Llama-2 and in Table~\ref{appendix:tab-llama3-all-cpt-strategies} we show performance for Llama-3.1 model. In case of Llama-2 the best pretraining strategy was Two-Stage and and for Llama-3.1 was End-To-End.

\paragraph{Mistral and Qwen results.} We present the Mistral results in Table~\ref{appendix:tab-mistral-all-cpt-strategies} and Qwen results in Table~\ref{appendix:tab-qwen2-all-cpt-strategies}. 

In case of Qwen-2, we observe that at least one vocabulary adaptation strategies improve over BASE and CPT-Only in 7 out of 12 comparisons considered in the main text (\textit{Difficult\textsubscript{SD}}, \textit{Difficult\textsubscript{RS}}, \textit{Novel\textsubscript{RS}}).  Here, we observe a similar trend with vocabulary adaptation not helping in case of low OOV and novelty concentration (specifically in BioASQ-$\mathcal{S}$) as we observed in main text (RQ6). The best performing continual pretraining strategy was End-to-End.

In case of Mistral, we observe that none of the vocabulary adaptation strategies improved over BASE and CPT-Only in one out of 12 comparisons considered in the main text (\textit{Difficult\textsubscript{SD}}, \textit{Difficult\textsubscript{RS}}, \textit{Novel\textsubscript{RS}}). Here, we observe a similar trend with vocabulary adaptation not helping in case of low OOV and novelty concentration (specifically in BioASQ-$\mathcal{S}$) as we observed in main text (RQ6). The best performing continual pretraining strategy was Two-stage for all the datasets except EBM.

\paragraph{Error Analysis for Performance Gap in BioASQ-$\mathcal{M}$.} We split the test set into two subsets: (i) Subset-1: instances where BASE has better Rouge-LCS than ScafFix (512 data points), and (ii) Subset-2: instances where ScafFix has higher Rouge-LCS than BASE (444 data points).  We analyzed characteristics like fraction of Medical OOV words in Reference Summary (i.e., Difficult-RS Concentration), fraction of novel unigrams in reference summary (i.e., Novel-RS Concentration) and content overlap (measured by the standard metric Rouge-LCS) between source document and reference summary (i.e., Rouge-LCS overlap between source and reference). The differences between the two subsets in terms of these characteristics are compared in Table~\ref{appendix:tab-character_eval_BioASQ-M}. 

In terms of understanding the error, we made a key observation. The instances where BASE has better Rouge-LCS than ScafFix (Subset-1) have lesser values of Difficult-RS Concentration as well as Novel-RS Concentration, but higher values of Rouge-LCS overlap compared to Subset-2. Thus, for the instances that are less novel, have less OOV concentration, and are easier to infer, ScafFix is less helpful. This result from error analysis is also in line with the comparison of fine-grained settings (Table ~\ref{tab:results-settings}) reported in the main text, where we see general improvements for both Llama-2 and Llama-3.1 models for instances with higher OOV and higher novelty settings.

\section{Human Evaluation}\label{appendix:human-eval-app}
We conducted our survey on Prolific platform\footnote{https://www.prolific.com/} where we hired 9 medical experts from the platform across globe. All the annotators were shown 10 random samples from a pool of 30 summaries where the order of summaries was randomized and blinded (annotators have no idea which summary came from which model). The median time to complete was set at 30 mins and the annotators were paid at the rate of 8 UK pounds per hour based on the amount of time they took. We did not collect any PII from the participants explicitly other than what was provided by the platform. The task was conducted using Google Forms, with participants being shown a consent notice beforehand. The results are shown in Figure~\ref{fig:human-eval}.

\paragraph{Participation Criteria.} The filtering criteria for participants were kept same as that of MEDVOC~\cite{balde2024ijcai}:
\begin{enumerate}
    \item Age: $\ge 25$,
    \item Primary Language: English,
    \item Highest education level completed: Graduate degree (MA/MSc/MPhil/other), Doctorate degree (PhD/other), and
    \item Subject: Medicine, Health and Medicine, Biomedical Sciences.
\end{enumerate}

\paragraph{Annotation Guidelines.} The annotations were assessed across three key dimensions as outlined by ~\citet{fabbri-etal-2021-summeval}: \textbf{Coherence}, \textbf{Relevance}, and \textbf{Factual Consistency}. 

\textbf{\underline{Coherence}} evaluates the structural integrity of the summaries, focusing on whether the sentences are logically connected and contextually aligned. \textbf{\underline{Relevance}} measures the informativeness of the summaries, considering the provided query and the context source document to judge the relevance. \textbf{\underline{Faithfulness}} examines the accuracy of the stated facts, figures, and numerical data within the summaries, ensuring they can be directly verified against the input source. Notably, even if a summary presents accurate information, it is considered factually inconsistent if the claims cannot be substantiated solely by the given input. Relevance and Coherence are judged on a scale of 1-5; whereas factual consistency is binary (Yes/No).

To make the guidelines clear, we provide few examples of positive and negative examples to make people aware of what is a relevant, coherent, and factually consistent document vs what is not. The pdf is present in the github codebase.

\clearpage
\section{Representative Examples} \label{appendix:sec-rep-examples-scaffix}
We present examples from PubMedQA and EBM datasets where vocabulary adaptation strategy ScafFix helped improve the performance over BASE in higher OOV concentration settings from Llama-2 models.

\begin{table}[!ht]
    \centering
    \scriptsize
    \begin{tabular}{p{0.48\textwidth}}
        \hline
        \multicolumn{1}{c}{\textbf{Input}}\\ \hline
        \textbf{Query}: What medication best prevents migraine in children? \\
        \textbf{PubMed Abstract}: In a prospective study we compared propranolol, placebo, and self-hypnosis in the treatment of juvenile classic migraine. Children aged 6 to 12 years with classic migraine who had no previous specific treatment were randomized into propranolol (at 3 mg/kg/d) or placebo groups for a 3-month period and then crossed over for 3 months. After this 6-month period, each child was taught self-hypnosis and used it for 3 months. Twenty-eight patients completed the entire study. The mean number of headaches per child for 3 months during the placebo period was 13.3 compared with 14.9 during the propranolol period and 5.8 during the self-hypnosis period. Statistical analysis showed a significant association between decrease in headache frequency and self-hypnosis training (P = .045). There was no significant change in subjective or objective measures of headache severity with either therapy. \\ \hline

        \multicolumn{1}{c}{\textbf{Reference Summary} (OOV Concentration: $17.72\%$)}\\ \hline
        A comparative randomized controlled trial with multiple crossovers involving 33 children found that a self-hypnosis placebo decreased mean headache frequency from 13.3 per 3-month interval to 5.8 (P=.045), but found propranolol no different than placebo. Propranolol was also studied in a 3-armed randomized controlled trial in comparison with flunarizine-a drug likely to be efficacious and placebo. \\ \hline

        \multicolumn{1}{c}{\textbf{BASE Summary} (Rouge-L: $24.10$)}\\ \hline
        Self-hypnosis was associated with a significant decrease in headache frequency in a single randomized controlled trial of 28 children. \\ \hline

        \multicolumn{1}{c}{\textbf{ScafFix Summary} (Rouge-L: $40.00$)}\\ \hline
        In a single randomized controlled trial with crossover design in 28 children, self-hypnosis reduced headache frequency from a mean of 13.3 attacks/month to 5.8 attacks/month vs. no change for placebo (P =.045). \\ \hline
        
    \end{tabular}
    \caption{Example from EBM dataset comparing summary generated using vocabulary adaptation method ScafFix with BASE. }
    \label{appendix:tab-EBM-example}
\end{table}

\begin{table}[!ht]
    \centering
    \scriptsize
    \begin{tabular}{p{0.48\textwidth}}
        \hline
        \multicolumn{1}{c}{\textbf{Input}}\\ \hline
        \textbf{Query}: Should serum pancreatic lipase replace serum amylase as a biomarker of acute pancreatitis? \\
        \textbf{PubMed Abstract}: Serum pancreatic lipase may improve the diagnosis of pancreatitis compared to serum amylase. Both enzymes have been measured simultaneously at our hospital allowing for a comparison of their diagnostic accuracy. Seventeen thousand five hundred and thirty-one measurements of either serum amylase and or serum pancreatic lipase were made ... One thousand eight hundred and twenty-five patients had either elevated serum amylase and or serum pancreatic lipase. The medical records coded for pancreatitis in a further 55 whose enzymes were not elevated. Three hundred and twenty of these had radiological evidence of acute pancreatitis. Receiver operator characteristic analysis of the initial sample from patients received in the Emergency Department showed improved diagnostic accuracy for serum pancreatic lipase (area under the curve (AUC) 0.948) compared with serum amylase (AUC, 0.906, P<0.05) ... \\ \hline

        \multicolumn{1}{c}{\textbf{Reference Summary} (OOV Concentration: $47.09\%$)}\\ \hline
        It is concluded that serum pancreatic lipase is a more accurate biomarker of acute pancreatitis than serum amylase. \\ \hline

        \multicolumn{1}{c}{\textbf{BASE Summary} (Rouge-L: $29.17$)}\\ \hline
        Serum pancreatic lipase may improve the diagnosis of pancreatitis compared to serum amylase. Both enzymes have been measured simultaneously at our hospital allowing for a comparison of their diagnostic accuracy. \\ \hline

        \multicolumn{1}{c}{\textbf{ScafFix Summary} (Rouge-L: $76.47$)}\\ \hline
        Serum pancreatic lipase is a more sensitive and specific biomarker of acute pancreatitis than serum amylase. \\ \hline
        
    \end{tabular}
    \caption{Example from PubMedQA dataset comparing summary generated using vocabulary adaptation method ScafFix with BASE. Interestingly, BASE model suffered from LEAD-bias~\cite{see-etal-2017-get} (picking sentences from the front of the document as the summary) and missed the biomarker word completely.}
    \label{appendix:tab-PMQA-example}
\end{table}

\end{document}